\documentclass[manuscript,screen]{acmart}

\usepackage{amsmath,amsfonts,bm, bbm}

\def\eqref#1{equation~\ref{#1}}

\def\1{\bm{1}}

\def\vb{{\bm{b}}}
\def\vc{{\bm{c}}}

\def\vf{{\bm{f}}}
\def\vg{{\bm{g}}}
\def\vh{{\bm{h}}}
\def\vi{{\bm{i}}}

\def\vn{{\bm{n}}}
\def\vo{{\bm{o}}}

\def\vq{{\bm{q}}}
\def\vr{{\bm{r}}}

\def\vv{{\bm{v}}}

\def\vx{{\bm{x}}}

\def\vz{{\bm{z}}}

\def\mQ{{\bm{Q}}}

\def\mU{{\bm{U}}}
\def\mV{{\bm{V}}}
\def\mW{{\bm{W}}}

\DeclareMathAlphabet{\mathsfit}{\encodingdefault}{\sfdefault}{m}{sl}
\SetMathAlphabet{\mathsfit}{bold}{\encodingdefault}{\sfdefault}{bx}{n}

\usepackage{mdframed}

\usepackage{accents}
\newlength{\dhatheight}

\usepackage{adjustbox}
\usepackage{tikz}
\usepackage{enumitem}
\usetikzlibrary{er,positioning}

\definecolor{ao}{rgb}{0.0, 0.5, 0.0}
\definecolor{bblue}{rgb}{0.19, 0.55, 0.91}
\definecolor{bt}{rgb}{1.0, 0.44, 0.37}
\definecolor{blush}{rgb}{0.87, 0.36, 0.51}

\usetikzlibrary{fit,positioning}
\tikzstyle{block} = [rectangle, draw, fill=blue!20, 
    text width=12.8em, text centered, rounded corners, minimum height=4em]
\tikzstyle{line} = [draw, -latex']

\tikzstyle{cloud} = [draw, ellipse,fill=red!20, node distance=3cm,
    minimum height=2em]
\AtBeginDocument{%
  \providecommand\BibTeX{{%
    \normalfont B\kern-0.5em{\scshape i\kern-0.25em b}\kern-0.8em\TeX}}}

\setcopyright{acmcopyright}
\copyrightyear{2022}
\acmYear{2022}
\acmDOI{XXXXXXX.XXXXXXX}

\acmConference[FAccT '22]{Make sure to enter the correct
  conference title from your rights confirmation emai}{June 21--24,
  2021}{South Korea}
\acmBooktitle{FAccT 2022: ACM Conference on Fairness, Accountability, and Transparency ,
 June 21--24, 2022, Seoul, South Korea} 

\begin{document}

\title{NeuroView-RNN: It's About Time}

\author{CJ Barberan}
\affiliation{%
  \institution{Rice University}
  \city{Houston}
  \state{Texas}
  \country{USA}
  \postcode{77005}
}
\email{cb30@rice.edu}

\author{Sina Alemohammad}
\affiliation{%
  \institution{Rice University}
  \city{Houston}
  \state{Texas}
  \country{USA}
  \postcode{77005}
}
\email{sa86@rice.edu}

\author{Naiming Liu}
\affiliation{%
  \institution{Rice University}
  \city{Houston}
  \state{Texas}
  \country{USA}
  \postcode{77005}
}
\email{nl35@rice.edu}

\author{Randall Balestriero}
\affiliation{%
  \institution{Rice University}
  \city{Houston}
  \state{Texas}
  \country{USA}
  \postcode{77005}
}
\email{randallbalestriero@gmail.com}

\author{Richard G. Baraniuk}
\affiliation{%
  \institution{Rice University}
  \city{Houston}
  \state{Texas}
  \country{USA}
  \postcode{77005}
}
\email{richb@rice.edu}




\renewcommand{\shortauthors}{}

\begin{abstract}
Recurrent Neural Networks (RNNs) are important tools for processing sequential data such as time-series or video. Interpretability is defined as the ability to be understood by a person and is different from explainability, which is the ability to be explained in a mathematical formulation.
A key interpretability issue with RNNs is that it is not clear how each hidden state per time step contributes to the decision-making process in a quantitative manner. 
We propose \emph{NeuroView-RNN} as a family of new RNN architectures that explains how all the time steps are used for the decision-making process.
Each member of the family is derived from a standard RNN architecture by concatenation of the hidden steps into a global linear classifier. The global linear classifier has all the hidden states as the input, so the weights of the classifier have a linear mapping to the hidden states. Hence, from the weights, NeuroView-RNN can quantify how important each time step is to a particular decision. As a bonus, NeuroView-RNN also offers higher accuracy in many cases compared to the RNNs and their variants.
We showcase the benefits of NeuroView-RNN by evaluating on a multitude of diverse time-series datasets.

\end{abstract}




\maketitle

\section{Introduction}
Recurrent neural networks (RNNs) \cite{hochreiter1997long} are ubiquitous in deep learning, because their design enables them to process arbitrary length sequential data. For example, RNNs and their variants like the Gated Recurrent Unit (GRU) \cite{chung2014empirical} and Long Short-Term Memory (LSTM) \cite{hochreiter1997long} have been core components in numerous applications, such as machine translation \cite{cho2014learning}, image/video captioning \cite{wang2016image,song2018deterministic}, and action recognition \cite{gammulle2017two,ma2018time}. There are other works in studying the dynamics of generalization, learning, and initializations of RNNs using neural tangent kernels \cite{alemohammad2020recurrent,alemohammad2020scalable}. However, even though RNNs are powerful tools, they are challenging to interpret and explain.

From \cite{gilpin2018explaining}, they state that interpretability has many definitions. The definition of interpretability \cite{gilpin2018explaining} that we will use is the ability to be understood by a person. In addition, the term, explainability is different from interpretability and will be defined as the ability to be explained in a mathematical formulation.

Deep learning interpretability has become an important topic since RNNs and their variants have become useful in several applications. The minimal amount of interpretability and explainability does not reveal why they have this performance on their tasks. With more interpretability with these models, it can explain why they are performing well on their respective tasks. Plus, if the model is not performing well, it is difficult to assess which specific parts of the signal contribute to the final decision. Since the model is opaque and hard to interpret, this creates a black-box effort for a practitioner.

There are works that have contributed to providing interpretability/explainability about how these RNNs perform. For instance, the work from \cite{guo2019exploring} explores the connection of the input variables to the RNN in order to interpret the performance. Other works such as \cite{li2017recurrent,guan2019towards,jiang2020cold} focus on the interpretability of RNNs within their respective task, hence limiting the ability to interpret other tasks. \cite{jiang2020cold,foerster2017input,ming2019interpretable} create their own type of RNN in order to provide interpretability in their application. The works of \cite{guan2019towards,li2017recurrent,arras2019explaining} use some metrics to provide interpretability with the RNN/LSTM that they use for their application.

With the works that created an interpretable RNN (\cite{jiang2020cold,foerster2017input,ming2019interpretable}), the main issue is that the definition of interpretability is focused on their application, which would make it difficult to adapt to another application for interpretability. For instance, \cite{jiang2020cold} creates a finite-automaton RNN for text classification. This work would be hard to adapt to other applications. Another issue is that with the works mentioned, they cannot provide a mathematical formulation that explains the prediction with the hidden states.

\textbf{Contributions.} 
We propose NeuroView-RNN (NV-RNN) as a novel general framework that provides enhanced interpretability and explainability to classification.

\begin{enumerate}
    \item We introduce the NV-RNN framework, which consists of a concatenation of all of the hidden states to be the input to the linear classifier. This allows a linear mapping to all of the hidden states to the classes. This linear mapping allows interpretability and explainability since the prediction for each class is in a dot product of the weights from the linear classifier with the hidden states.
    
    \item Influenced from the work of \cite{barberan2021neuroview}, we provide interpretability and explainability defined in the Introduction to showcase how NV-RNN, NV-GRU, and NV-LSTM can provide more understanding especially with applications and architectures of RNNs, GRUs, and LSTMs. This is in stark contrast to the work of \cite{barberan2021neuroview} that used convolutional neural networks (CNNs) for image classification which cannot on its own be used in specific RNN architectures like bidirectional RNNs or variable length input.
    
    \item We have better performance on most of the datasets from Table~\ref{table:1} to show why this NV-RNN framework should be used. In addition, we have on par performance on some of the datasets from Table~\ref{table:1} and \ref{table:2}. The results show that we have the accuracy performance compared to typical RNNs and are able to interpret and explain the performance.
    
    \item We use numerous case studies about weight initialization, bidirectional GRUs, semantic analysis, video action recognition, and counterfactuals to show the benefit of NV-RNN.
\end{enumerate}



\section{Background}

\subsection{Recurrent Neural Networks}

Given an input sequence data
$\vx = \{\vx_t\}_{t = 1}^{T}$ of length $T$ with data at time $t$, $\vx_t\in\mathbb{R}^m$, an RNN performs the following recursive block computation at each time step $t$
\begin{align}
  \vh^{\left(t\right)}(\vx) &= \mathcal{F}_{\theta}(\vh^{\left(t-1\right)}(\vx), \vx_t) \in \mathbb{R}^n \label{rnn-main},
\end{align}
where $h^{(0)}(\vx) = 0$ and $n$ is the number of parameters for the hidden state. $\mathcal{F}_{\theta} : \mathbb{R}^n\times \mathbb{R}^m \rightarrow \mathbb{R}^n$ is the hidden time steps mapping with time-agnostic parameters $\theta$. Each recurrent architecture has a different $\mathcal{F}_{\theta}$. For a simple RNN \cite{ElmanRNN} we have this formulation,
\begin{align}
    \mathcal{F}_{\theta}(\vh^{\left(t-1\right)}(\vx), \vx_t) = \vh^{(t)}(\vx) = \phi \left( \mW\vh^{\left(t-1\right)}(\vx) + \mU\vx_t + \vb \right),
\end{align}
where $\phi: \mathbb{R} \rightarrow \mathbb{R}$ is the activation function that act point-wise on a vector and $\theta = \mathrm{vect}\big[\{\mW,\mU,\vb \}\big]$ contains the mapping parameters. In case of simple RNNs, we use the sigmoid function $\phi(\alpha) = \frac{1}{1 + e^{( - \alpha)}}$. Other RNNs variants such as GRU \cite{cho2014learning} and LSTM \cite{hochreiter1997long} have a more complex mapping $\mathcal{F}_{\theta}$. See Supplementary Material Section \ref{sm-arch} for more details. 

Bidirectional recurrent architectures \cite{schuster1997bidirectional}, use two separated \emph{forward} and \emph{reverse} direction
\begin{align}
    \vh_f^{\left(t\right)}(\vx) &= \mathcal{F}_{\theta_f}(\vh_f^{\left(t-1\right)}(\vx), \vx_t) \\
    \vh_r^{\left(t\right)}(\vx) &= \mathcal{F}_{\theta_r}(\vh_r^{\left(t+1\right)}(\vx), \vx_t), 
\end{align}
where $\theta_f$ and $\theta_r$ are independent of each other and together form the network parameters $\theta = \{\theta_{f},\theta_r \}$. The final hidden state is obtained by concatenation of each direction hidden states,
\begin{align}
    \vh^{(t)}(\vx)= 
  \left[\vh_f^{(t)}(\vx)^\top,\vh_r^{(t)}(\vx)^\top\right]^\top \in \mathbb{R}^{2n}. \label{eq:bidirectional}
\end{align}

The output of a \emph{many to one} recurrent architecture is generally a linear transform of the last hidden state $T$:
\begin{align}
    f_{\theta}(\vx) = \mV\vh^{(T)}(\vx) \in \mathbb{R}^d.
\end{align}

Many-to-one recurrent architecture refers to when the input is a sequence of data but the output is decided at the end. They are used in applications like sentiment analysis or time-series classification when the input is a sequence and the output is to decide which class it is.
For recurrent architectures with average pooling \cite{shen2016empirical}, the output is a linear transform of the sum of all hidden states:
\begin{align}
    f_{\theta}(\vx) = \sum_{t = 1}^{t} \mV \vh^{(t)}(\vx) \in \mathbb{R}^d.
\end{align}


\section{NV-RNN: Interpretable and Explainable Recurrent Neural Network}

NV-RNN is inspired by the work in \cite{barberan2021neuroview} except in that work the authors only focused on 2D CNNs. The work in \cite{barberan2021neuroview} focuses on the spatial filters of CNNs whereas our work is adapted to use for RNNs. This paper focuses on RNNs with their variants and the hidden states per time. \cite{barberan2021neuroview} could only be used for CNN architectures and we adapt it to the RNN architecture which is different since CNNs focus on spatial features while RNNs focus on temporal features. In addition, the applications to RNNs have not been explored to specific RNN architectures like bidirectional RNN or varying input length which scenario is not present with CNNs and image classification. Therefore, adapting the work from \cite{barberan2021neuroview} is non-trivial and brings merit into interpretability and explainability in the definitions that we denoted in the introduction. 

\subsection{NV Architecture Description}

Given the sequences of hidden states calculated in Equation \ref{rnn-main} for any recurrent network, the output from NV-RNN is obtained first by acquiring the hidden states for all time steps
\begin{align}
\label{quntized}
    \vq^{(t)}(\vx) = \mathrm{ReLU} \left(\vh^{(t)}(\vx) \right), 
\end{align}
where $\mathrm{ReLU}(\alpha) = \max (\alpha, 0)$. In the next step, the output is calculated using
\begin{align}
    f_{\theta}(\vx) = \sum_{t = 1}^{T} \left(\mV^{(t)}\right)^\top\vq^{(t)}(\vx) =  \sum_{t = 1}^{T} f^{(t)}(\vx) \in \mathbb{R}^{d}. \label{eq:vq}
\end{align}

This is equivalent to concatenate all the hidden states per time step into a large hidden state vector $\mQ(\vx)$, where 
\begin{align}
   \mQ(\vx)=
  \left[\vq^{(1)}(\vx)^\top,\vq^{(2)}(\vx)^\top,\dots,\vq^{(T)}(\vx)^\top\right]^\top \in \mathbb{R}^{nT},
\end{align}
and concatenate all linear output weights into one large matrix
\begin{align}
   \mV=
  \left[\mV^{(1)},\mV^{(2)},\dots,\mV^{(T)}\right] \in \mathbb{R}^{nT \times d}.
\end{align}
and calculate the output as the following
\begin{align}
    f_{\theta}(\vx) = \mV^\top \mQ(\vx) \in \mathbb{R}^d. \label{eq:VtQ}
\end{align}

Where unlike commonly used recurrent architectures, the NV-RNN concatenates all of the hidden states as the input to the linear classifier. This is different from a typical RNN where the last hidden state is the input to the linear classifier. This concatenation does increase the size of the input but is needed for both interpretability/explainability and the performance in Table~\ref{table:1}. 

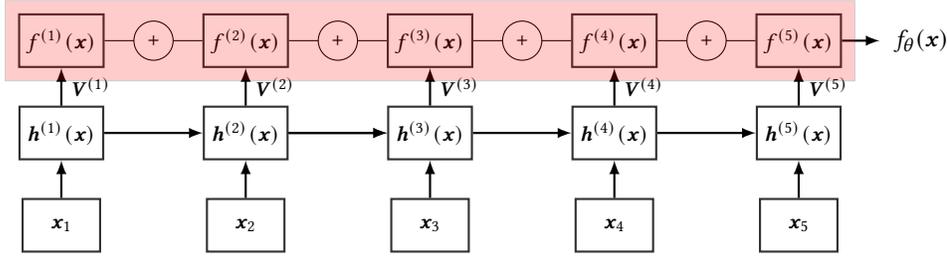
\begin{figure}
\centering
\begin{tikzpicture}[object/.style={thin,double,<->}]
    \tikzstyle{main}=[rectangle, minimum width = 10.3mm, minimum height = 7mm, thick, draw =black!80, node distance = 5mm]
    \tikzstyle{main2}=[circle, minimum size = 10mm, thick, draw =black!80, node distance = 5mm]
    \tikzstyle{connect}=[-latex, thick]
    \tikzstyle{box}=[rectangle, draw=black!100]

    \node[main] (old4) [] {\small$f^{(2)}(\vx)$};
    \node[main] (old1) [left=1.3cm of old4] {\small $f^{(1)}(\vx)$};
    \node[main] (old7) [right=1.3cm of old4] {\small $f^{(3)}(\vx)$};
    \node[main] (old5) [below=0.5cm of old4] {\small $\vh^{(2)}(\vx)$};
    \node[main] (old2) [below=0.5cm of old1] {\small $\vh^{(1)}(\vx)$};
    \node[main] (old8) [below=0.5cm of old7] {\small $\vh^{(3)}(\vx)$};
    \node[main] (new1) [right=1.3cm of old7] {\small $f^{(4)}(\vx)$};
    \node[main] (new2) [below=0.5cm of new1] {\small $\vh^{(4)}(\vx)$};
    \node[main] (new3) [below=0.5cm of new2] {\small $\vx_4$};
    \node[main] (new4) [right=1.3cm of new1] {\small $f^{(5)}(\vx)$};
    \node[main] (new5) [below=0.5cm of new4] {\small $\vh^{(5)}(\vx)$};
    \node[main] (new6) [below=0.5cm of new5] {\small $\vx_5$};
    \node[main] (xtm) [below=0.5cm of old2] {\small $\vx_{1}$};
    \node[main] (xt)  [below=0.5cm of old5] {\small $\vx_{2}$};
    \node[main] (xtp) [below=0.5cm of old8] {\small $\vx_{3}$};
    \coordinate[right=0.5cm of new4] (c1);
    \coordinate[right=1.2cm of new4] (c2);
    \node[text width=1cm] at (c2) {$f_{\theta}(\vx)$};
      
    \path (new1) edge node[draw,fill=white,circle]{+} (new4)
        (old1) edge node[draw,fill=white,circle]{+} (old4)
        (old4) edge node[draw,fill=white,circle]{+} (old7)
        (old7) edge node[draw,fill=white,circle]{+} (new1);

    \path 
        (old2) edge [connect] node[above] {}(old5)
        (old5) edge [connect] node[above] {}(old8)
		(old5) edge [connect] node[above] {} (old8)
		(old8) edge [connect] node[above] {} (new2)
		(new2) edge [connect] node[above] {} (new5)
		(new4) edge [connect] node[above] {} (c1)
		(old2) edge [connect] node[right] {\small$\mV^{(1)}$}(old1)
		(old5) edge [connect] node[right] {\small$\mV^{(2)}$}(old4)
		(old8) edge [connect] node[right] {\small$\mV^{(3)}$} (old7)
		(new2) edge [connect] node[right] {\small$\mV^{(4)}$} (new1)
		(new5) edge [connect] node[right] {\small$\mV^{(5)}$} (new4)
		;
	\node[rectangle, inner sep=1.7mm,draw=black!100, fit= (old1)(old7)(new4), fill=red,opacity=.2](t) {};

    \path (xtm) edge [connect] node[right] {}(old2)
    (xt) edge [connect] node[right] {}(old5)
    (xtp) edge [connect] node[right] {}(old8)
    (new3) edge [connect] node[right] {}(new2)
    (new6) edge [connect] node[right] {}(new5);

\end{tikzpicture}
\caption{Depiction of the NV-RNN framework. Every RNN can be converted to a NV model by having every hidden state concatenated to the linear classifier. The input to the linear classifier is the concatenation of all the hidden states. This provides a mapping to evaluate which time steps are the most relevant to each class. }
\label{fig:my_label}

\end{figure}

Figure~\ref{fig:my_label} depicts our NV-RNN where it displays how the hidden states are aggregated in order to provide a classification decision. Note that this network is applied to many-to-one applications.

The use of new weights for the linear classifier at each time step limits the NV-RNN to datasets where all data sequences have the same length. In the case where the data sequences have variable length, then zero padding will be used. This interpretability especially in terms of which time steps are resonating with the class helps explain which time steps are contributing the most. The entries of $\mV$ will state in a numeric manner how much each weight is contributing to classification.

\subsection{Interpretable and Explainable Linear Classifier}

Recall that the final input to the linear classifier is the concatenation of the hidden states from all the time steps. When training is complete, we look at the weights per class and since we have the linear relationship of the linear classifier to all the hidden states, we have this ordered mapping of the weights to the hidden states. This ordering is done by concatenating the first hidden state starting from the first time step and concluding with the final time step's hidden state.
 Hence, we inspect which of the hidden states per time are contributing to the decision-making process for every class.
 Recall that from Equation \ref{eq:VtQ} we have NV-RNN denoted in that form. Then we can substitute the right hand side to have in it in this new form (Equation~\ref{eq:vq}). The classification depends on the concatenation of all the hidden states, $\mQ$, and the weights from the linear classifier. For each class of the output we get
 
 \begin{align}
    \left[ f_{\theta}(\vx) \right]_i=\vv_i^\top \mQ(\vx) \in \mathbb{R}, \,\,\,\, \forall \,\, i \in [d], \label{eq:class}
\end{align}
where $\vv_i \in \mathbb{R}^{nT}$ is the $i$th row of $\mV$ that corresponds to class $i$. Since each class output is obtained by inner product of $\mQ(\vx)$ and $\vv_i$, each class acquires a set of distinct learned $\vv_i$ weights and we can interpret their values for each class. The goal is that in the training phase, the weights of $\mV$ are generalized well for a test set and we can inspect which time steps are contributing the most by looking at the values. Each row of $\mV$ corresponds to a class, so for every input that is sent, the softmax will choose the class with the highest score when each of the class's weights are multiplied by the time steps' hidden state activations.

In addition to interpretability, we provide explainability of the classification. For each class, their $\vv$ explain the classification since $\mQ(\vx)$ is constant for each class and the determining factor are each class's $\vv$ since the highest dot product from Equation~\ref{eq:class} will determine which class is chosen.
 
This interpretability and explainability can provide more understanding of the training dynamics of NV-RNN. This provides additional analysis to why some models perform better than others. In Figures~\ref{fig:chinatownNV}, \ref{fig:fungiNV}, \ref{fig:chlorineNV}, and \ref{fig:insectNV} we can see which time steps are critical based on their weight values from looking at each class's $\vv$.
 
\subsection{Experimental Setup}
We demonstrate the performance of NV-RNN, NV-GRU and NV-LSTM against other RNN, GRU, and LSTM architectures that are difficult to interpret/explain. The premise is that if we are on par or better than these models it is beneficial to use. The RNN models that are used besides RNN are GRU and LSTM. Also we monitor the effect of average pooling with RNNs, LSTMs, and GRUs in our comparison. The idea to benchmark against average pooling RNNs came from these works, \cite{wang2019comparison,kao2020comparison}, since they have better performance than the standalone RNNs, LSTMS, and GRUs. Datasets that were used are the UCR repository \cite{Dau_2019}, Large Movie Review \cite{maas-EtAl:2011:ACL-HLT2011}, and UCF11 \cite{liu2009recognizing}. Do note that these tasks fall under the many-to-one regime for RNN classification. The UCR datasets had the input of the same length while the Large Movie Review dataset has varying input length. For UCF11, we use a sampler that we sample 50 sequential frames from each video. For the UCR and Large Movie Review datasets, we choose the best accuracy from each model where we vary the hidden state size to 32, 64, and 128. For UCF11, we use the hiddens state size of 32. All of the hyperparameters for the experiments are detailed in the Supplementary Material Section in \ref{section:hyper}.

\begin{table}
\centering
\caption{The best performing NV-RNN, NV-GRU, and NV-LSTM models against the best performing RNN, GRU, and LSTM models. For most of the datasets, the NV-RNN, NV-GRU, and NV-LSTM models outperform especially on datasets that have more time steps.}
{
\begin{adjustbox}{width=\linewidth,center}
\begin{tabular}{|l|ccc|ccc|ccc|}
\toprule
\textbf{Data set}        & \textbf{RNN} & \textbf{GRU} &\textbf{LSTM} & \textbf{RNN-AVG} & \textbf{GRU-AVG} & \textbf{LSTM-AVG} & \textbf{NV-RNN} & \textbf{NV-GRU} & \textbf{NV-LSTM} \\ \hline 
Adiac  & 35.8\% & 37.08\% & 49.61\% & 10.23\% & 31.45\% & 16.87\% & 69.56\% & 68.28\% & \textbf{74.68\%} \\
BME & 88.66\% & 94.66\% & 80\% & 84\% & 84.66\% & 84.66\% & \textbf{99.3\%} & 98.66\%  & 98.66\% \\
CBF & 60.66\% & 94.55\% & 90\% & 97.33\% & 98.66\% & \textbf{99.77\%} & 97.77\% & 98.44\% & 98.55\% \\
Chinatown & 74.34\% & 97.37\% & 97.66\% & \textbf{98.83\%} & 98.25\% & 98.54\% & 97.95\% & 97.08\% & 98.54\% \\
Chlorine Concentration & 58.17\% & 60.1\% & 57.73\% & 55.39\% & 57.05\% & 55.88\% & \textbf{83.95\%} & 78.15\% & 72.39\% \\
Fungi & 49.46\% & 58.6\% & 68.81\% & 60.21\% & 58.6\% & 75.26\% & 96.77\% & 98.92\% & \textbf{99.46\%} \\
Ham & 69.52\% & 68.57\% & 69.52\% & 74.28\% & \textbf{81.9\%} & 80.95\% & 78.09\% & 80.95\% & 78.09\% \\
Haptics & 42.2\% & 41.55\% & 41.88\% & 33.76\% & 44.48\% & 41.88\% & \textbf{46.42\%} & 46.1\% & 45.77\% \\
Herring & 67.18\% & 67.18\% & 68.75\% & 67.18\% & 65.62\% & 68.75\% & 68.75\% & \textbf{73.43\%} & 68.75\% \\
Insect Regular & \textbf{100\%} & \textbf{100\%} & \textbf{100\%} & \textbf{100\%} & \textbf{100\%} & \textbf{100\%} & \textbf{100\%} & \textbf{100\%} & \textbf{100\%} \\
Insect Small & \textbf{100\%} & \textbf{100\%} & \textbf{100\%} & \textbf{100\%} & \textbf{100\%} & \textbf{100\%} & \textbf{100\%} & \textbf{100\%} & \textbf{100\%} \\
InsectWingbeat & 28.93\% & 49.34\% & 43.58\% & 28.48\% & 46.41\% & 39.94\% & 64.29\% & 64.54\% & 63.88\% \\
Meat & 48.33\% & 50\% & 50\% & 66.66\% & 86.66\% & 81.66\% & \textbf{96.66\%} & \textbf{96.66\%} & \textbf{96.66\%} \\
OliveOil & 46.66\% & 50\% & 40\% & 40\% & 80\% & 40\% & \textbf{93.33\%} & \textbf{93.33\%} & \textbf{93.33\%} \\
Plane & 89.52\% & 79.04\% & 95.23\% & 65.71\% & 98.09\% & 70.47\% & 99.04\% & \textbf{100\%} & 99.04\% \\
Rock & 64\% & 74\% & 68\% & 56\% & 62\% & 60\% & 80\% & 76\% & \textbf{82\%} \\
SmoothSubspace & 91.33\% & 89.33\% & 90.66\% & 90.66\% & 91.33\% & 86.66\% & 91.33\% & \textbf{96\%} & 94\% \\
Synthetic Control & \textbf{99.66\%} & 98.66\% & 98.33\% & 94.33\% & 95.66\% & 97.33\% & \textbf{99.66\%} & 99.3\% & 99.3\% \\
UMD & 74.3\% & 99.3\% & 86.8\% & 75\% & 92.36\% & 72.22\% & \textbf{100\%} & \textbf{100\%} & \textbf{100\%} \\
Wine & 59.25\% & 59.25\% & 62.96\% & 75.92\% & 79.62\% & 74.07\% & \textbf{100\%} & \textbf{100\%} & \textbf{100\%} \\

\bottomrule
\end{tabular}
\label{table:1}
\end{adjustbox}
}
\label{tab:real-data-svm}
\end{table}

\begin{table}
\centering
\caption{The best performing NV-GRU models against the best performing non NV-GRU models with the Large Movie Review dataset. The dataset has sentences of reviews of variable length. NV-GRU performs close to on par despite zero padding.}
{
\begin{adjustbox}{width=0.5\linewidth,center}
\begin{tabular}{|l|l|c|c|}
\toprule
\textbf{Data set}   & \textbf{Embedding}     & \textbf{GRU} & \textbf{NV-GRU} \\ \hline 
Large Movie Review & Word2Vec & \textbf{91.87\%} & 90.12\% \\
Large Movie Review & FastText & \textbf{91.46\%} & 89.56\% \\
Large Movie Review & GloVe & \textbf{89.98\%} & 87.76\% \\

\bottomrule
\end{tabular}
\label{table:2}
\end{adjustbox}
}
\end{table}

\begin{table}
\centering
\caption{The best performing NV-CNN-RNN, NV-CNN-GRU, and NV-CNN-LSTM models against the best performing CNN-RNN, CNN-GRU, and CNN-LSTM models. The dataset is UCF11 where the task is video action recognition. }
{
\begin{adjustbox}{width=0.9\linewidth,center}
\begin{tabular}{|l|c|c|c|c|c|c|}
\toprule
\textbf{Data set}   & \textbf{NV-CNN-GRU}     & \textbf{NV-CNN-RNN} & \textbf{NV-CNN-LSTM} & \textbf{CNN-GRU} & \textbf{CNN-RNN} & \textbf{CNN-LSTM} \\ \hline 
UCF11 & 76.4\% & 74.3\% & \textbf{78.3\%} & 69.3\% & 72.1\% & 72.4\% \\

\bottomrule
\end{tabular}
\label{table:3}
\end{adjustbox}
}
\end{table}

\subsection{Results}
We see from Table~\ref{table:1} that for most of the datasets, any of the NV-RNN, NV-GRU, and NV-LSTM models outperform the traditional RNN, GRU, and LSTM and its variants. For the exact accuracies among the different hidden state sizes look into the Supplementary Material Section \ref{section:ablation}. It is interesting to note that with the datasets that have a smaller amount of timesteps, the RNNs will perform better compared to their gating variants. In addition, with the Adiac dataset, the results show that using average pooling on the hidden states is not the optimal solution. This is why NV-RNN was developed to have the linear classifier learn which of the time steps should have higher positive or negative weight values for the different classes. From Table~\ref{table:2} we apply NV-GRU to a dataset where the input has varying length and we perform on par. The reason is that the maximum length we set is 1000 and any reviews under that maximum length will have zero padding. From Table~\ref{table:3}, we outperform on another application called video action recognition. Note, that in this application, we have to use CNN filter units and RNN hidden state units as input to the linear classifier.

With the great performance, there is a cost associated with it. The cost is that the input for the linear classifier greatly increases. The input of the linear classifier size greatly increases since we concatenate all the hidden states from the input. This is in stark contrast from the input being the last hidden state or the averaging of the hidden states. Yet, even with the additional memory overhead, we can still run these NV-RNN, NV-GRU, and NV-LSTM models with an 8 GB GPU.


\section{NV-RNN Time Analysis}

\subsection{Interpreting the Time Steps}

Now with the experimental results showing that the NV-RNN, NV-GRU, and NV-LSTM models outperform RNNs, GRUs, and LSTMs among multiple datasets, the next step is to explain which time steps are pivotal for the classification portion. This is something that is lacking with the models that we compared. The weights from the linear classifier have a linear mapping between the different hidden states per time in order to directly observe which hidden states are responsible for each class's decision. Each class will have its own distribution of linear classifier weights. In the following section, we will observe different classes' linear weights for different datasets, different hidden state and weight initializations, and other case studies.

\begin{figure}[h!]%
\centering
\includegraphics[width=.45\textwidth]{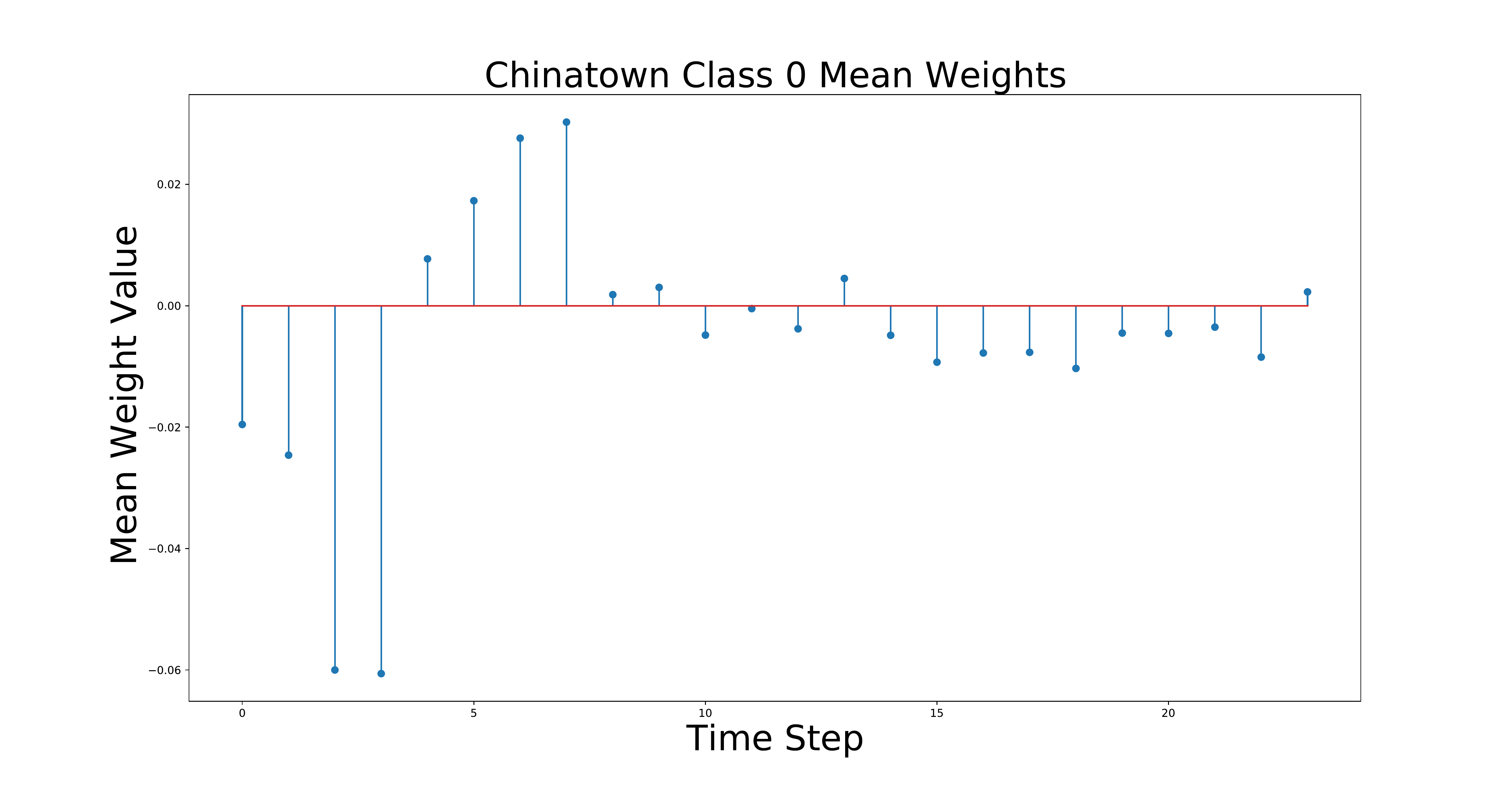}\hfill
\includegraphics[width=.45\textwidth]{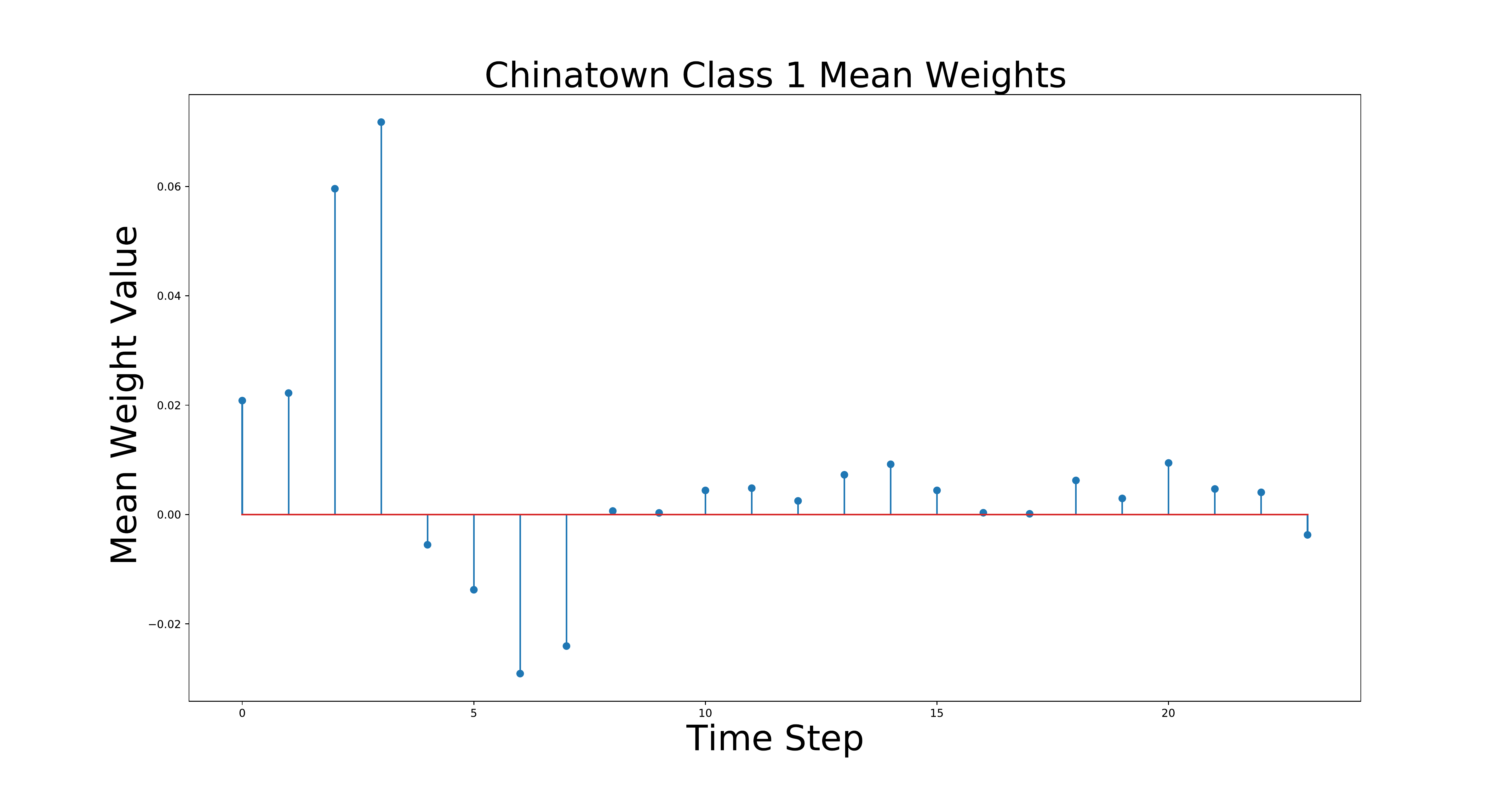}\hfill
    \caption{(Left) NV-GRU weights for class 0. (Right) NV-GRU weights for class 1. For the Chinatown dataset, there are two classes and the weights for each class are drastically different. Hence each class's prediction focuses on different time steps. In regard to class 0, the most important time steps are 4, 5, 6, and 7. In regard to class 1, the most important time steps are 0, 1, 2, and 3.}%
    \label{fig:chinatownNV}%
\end{figure}

\begin{figure}[h!]%
\centering
\includegraphics[width=.32\textwidth]{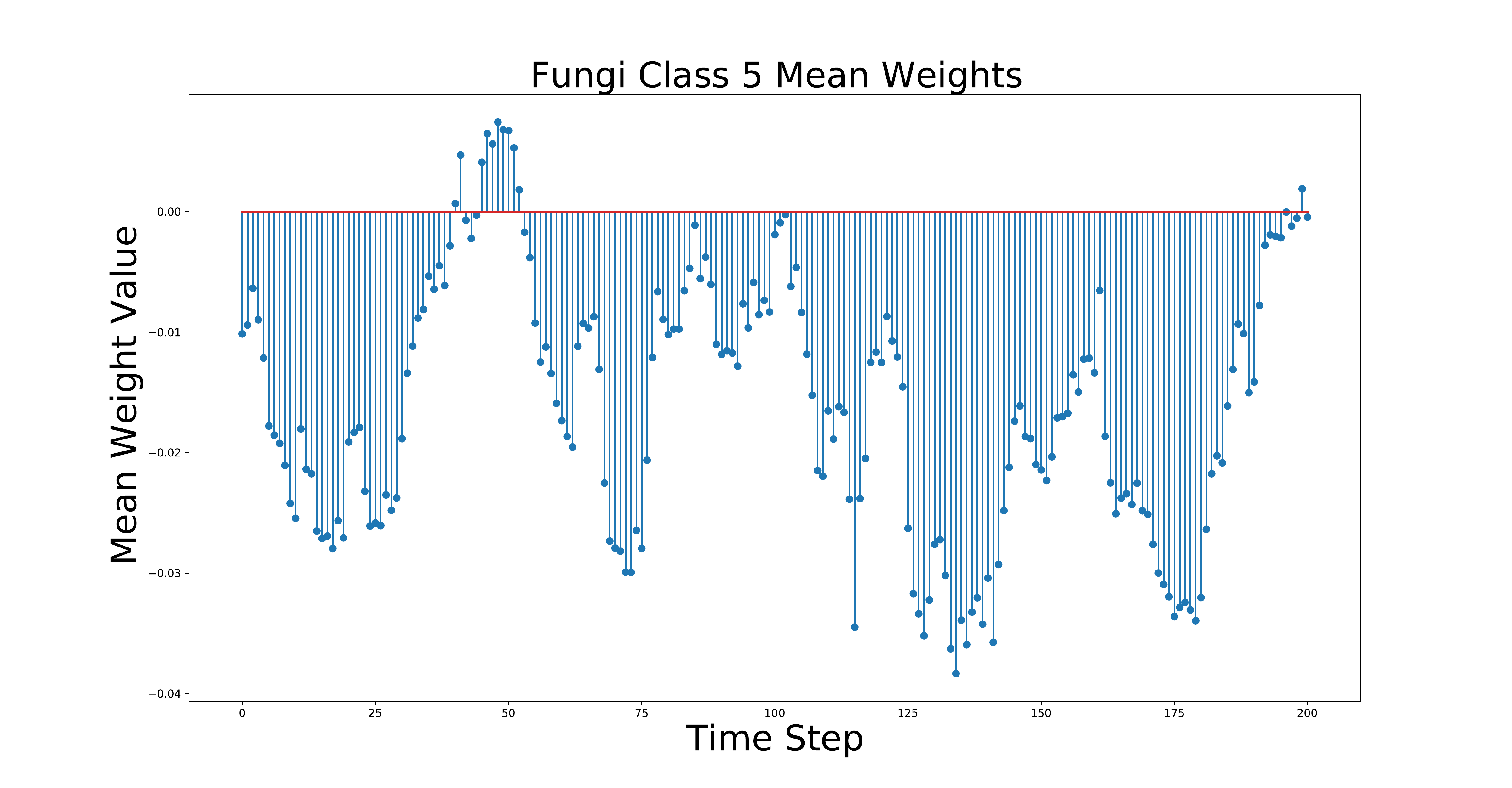}\hfill
\includegraphics[width=.32\textwidth]{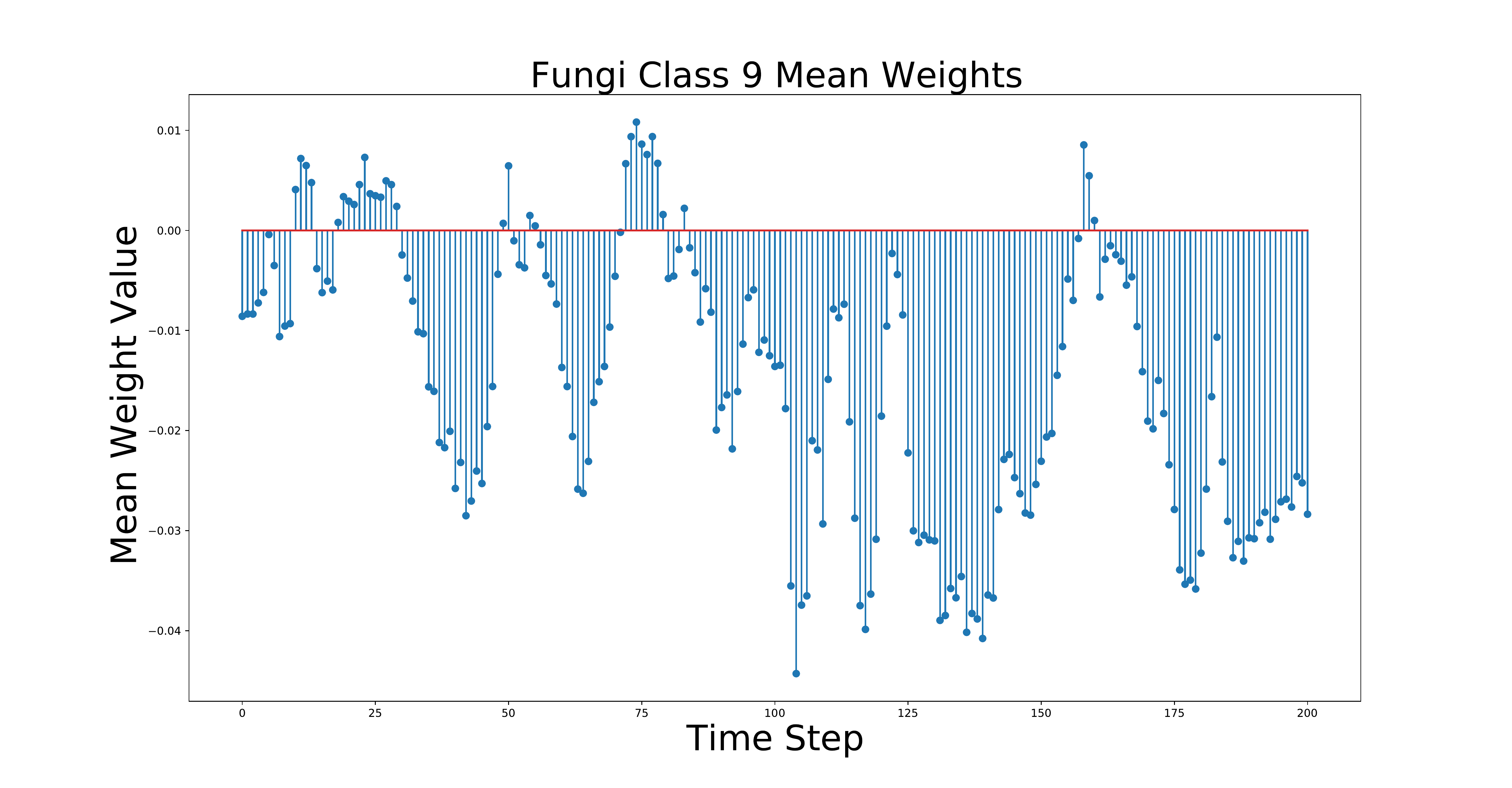}\hfill
\includegraphics[width=.32\textwidth]{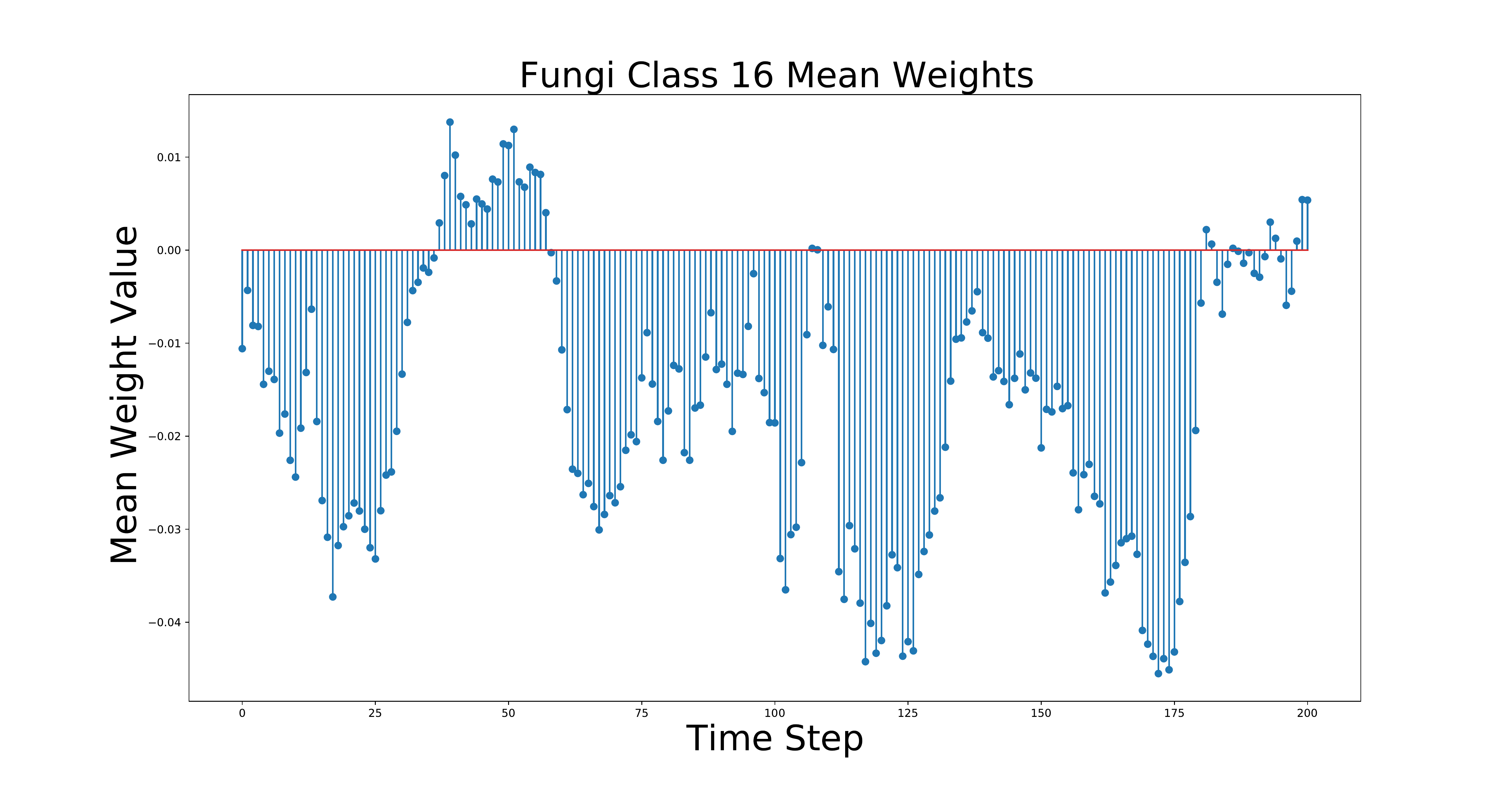}
    \caption{(Left) NV-GRU weights for class 5. (Middle) NV-GRU weights for class 9. (Right) NV-GRU weights for class 16. For the Fungi dataset, there are a total of 18 classes. Compared to class 5's weights, class 9's weights are one of the most different based on the cosine similarity of the weights for each of these classes. For class 16, it is one of the most similar classes to class 5 based on the cosine similarity of the classes. Inspecting the NV-GRU weights from the linear classifier can aid in interpreting which classes are similar or dissimilar to each other.}%
    \label{fig:fungiNV}%
\end{figure}

\begin{figure}[h!]%
\centering
\includegraphics[width=.32\textwidth]{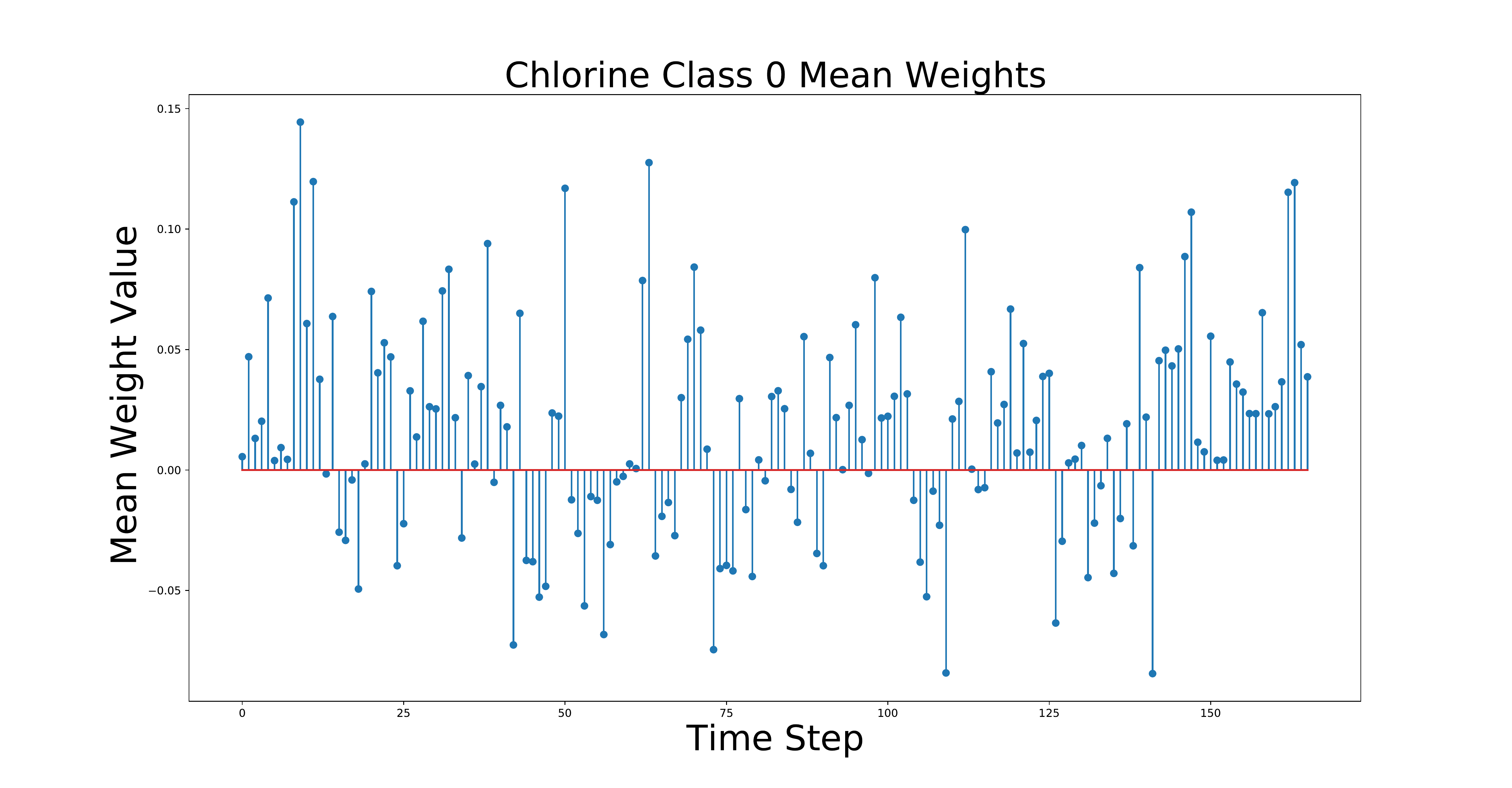}\hfill
\includegraphics[width=.32\textwidth]{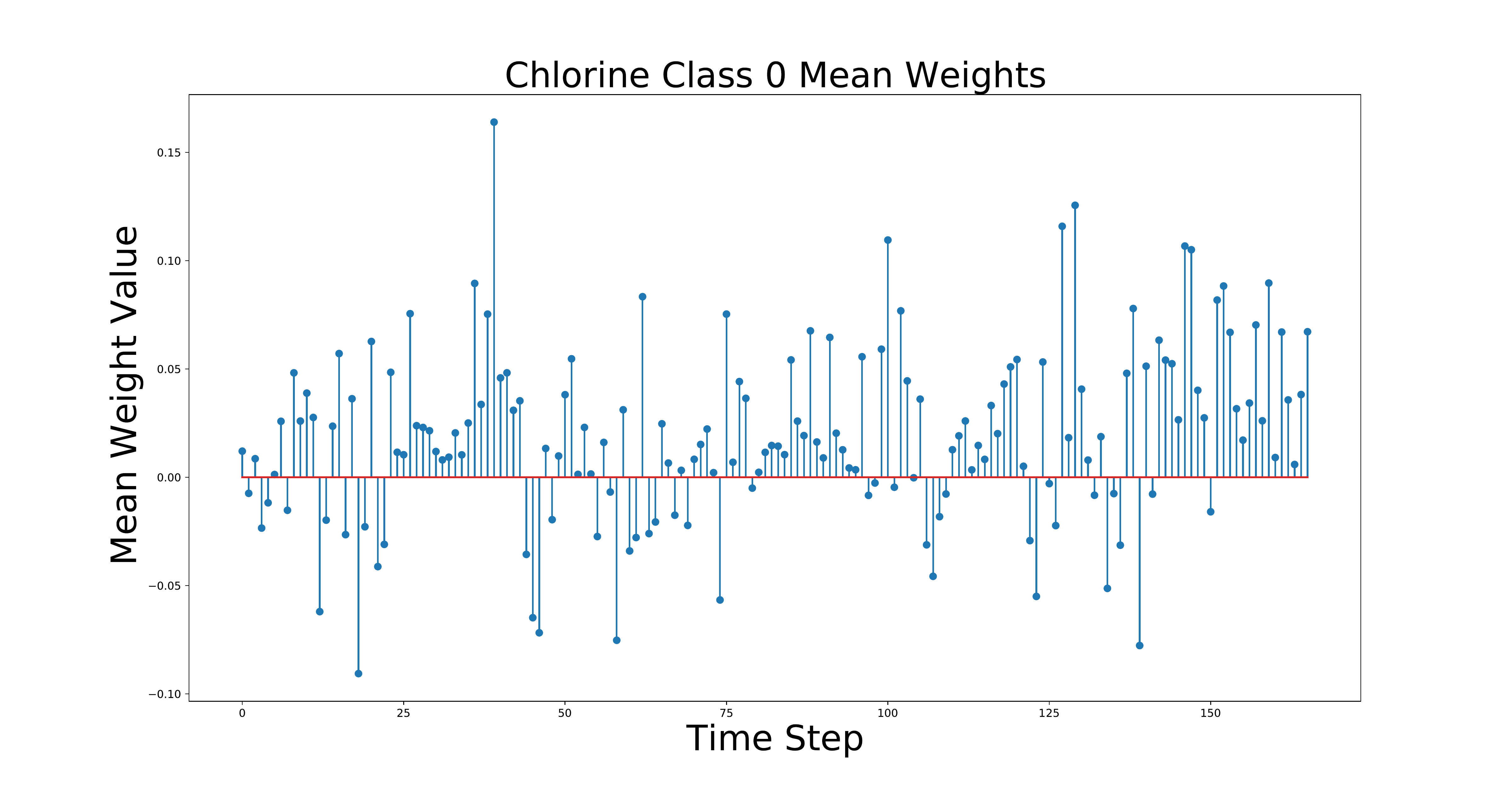}\hfill
\includegraphics[width=.32\textwidth]{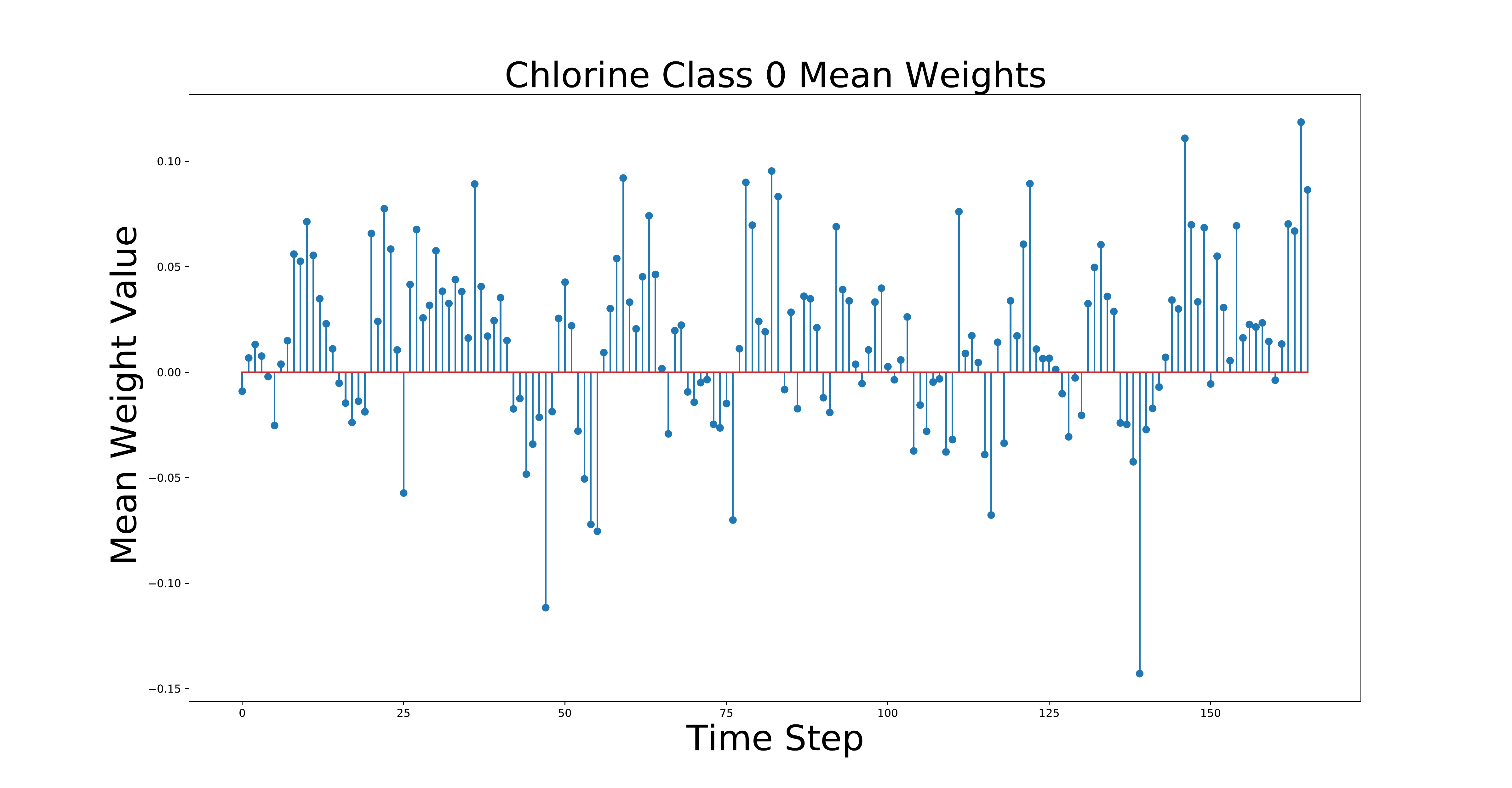}
    \caption{(Left) NV-GRU weights for class 0. (Middle) NV-RNN weights for class 0. (Right) NV-LSTM weights for class 0. From Table~\ref{table:1}, each of the NV models had different test accuracies and by inspecting each of the models, the prioritization of the time steps are different. The last time steps seem to have a similar trend for all of the NV models but the middle time steps have different prioritizations.}%
    \label{fig:chlorineNV}%
\end{figure}

This interpretation is lacking from RNNs and their variants since the input to the linear classifier is the last hidden state. Then for the RNNs with average pooling, the notion of averaging the hidden states does not produce the best results as shown. Hence, by having all the hidden states as the input for the linear classifier, we learn the weights for each class to prioritize all the time steps. Each class will have a set of learnable weights that can be different from other classes. Figures~\ref{fig:chinatownNV}, \ref{fig:fungiNV}, and \ref{fig:chlorineNV} show the weights from different NV-RNN models for different classes.

\begin{figure}[h!]%
\centering
\includegraphics[width=.45\textwidth]{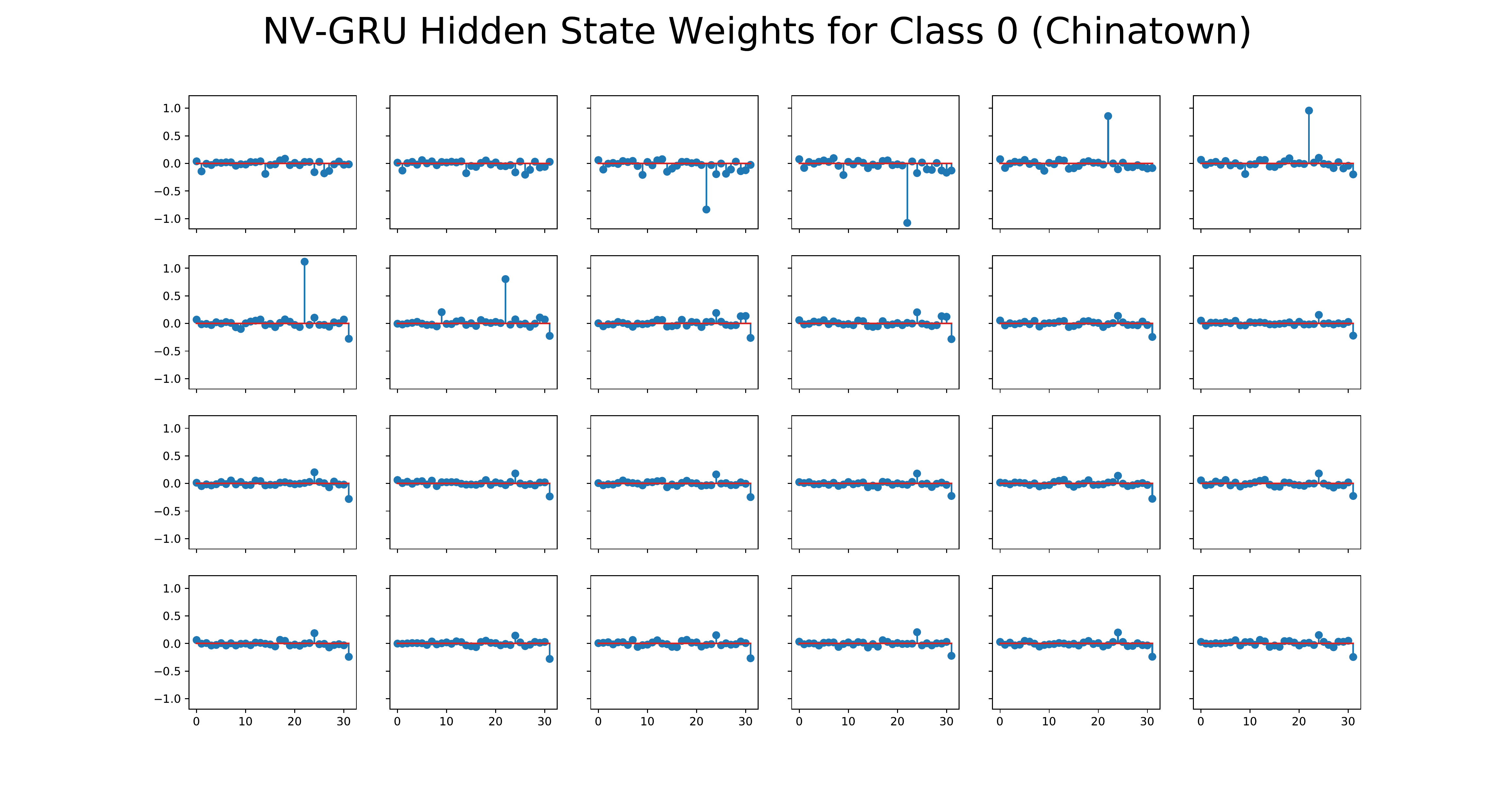}\hfill
\includegraphics[width=.45\textwidth]{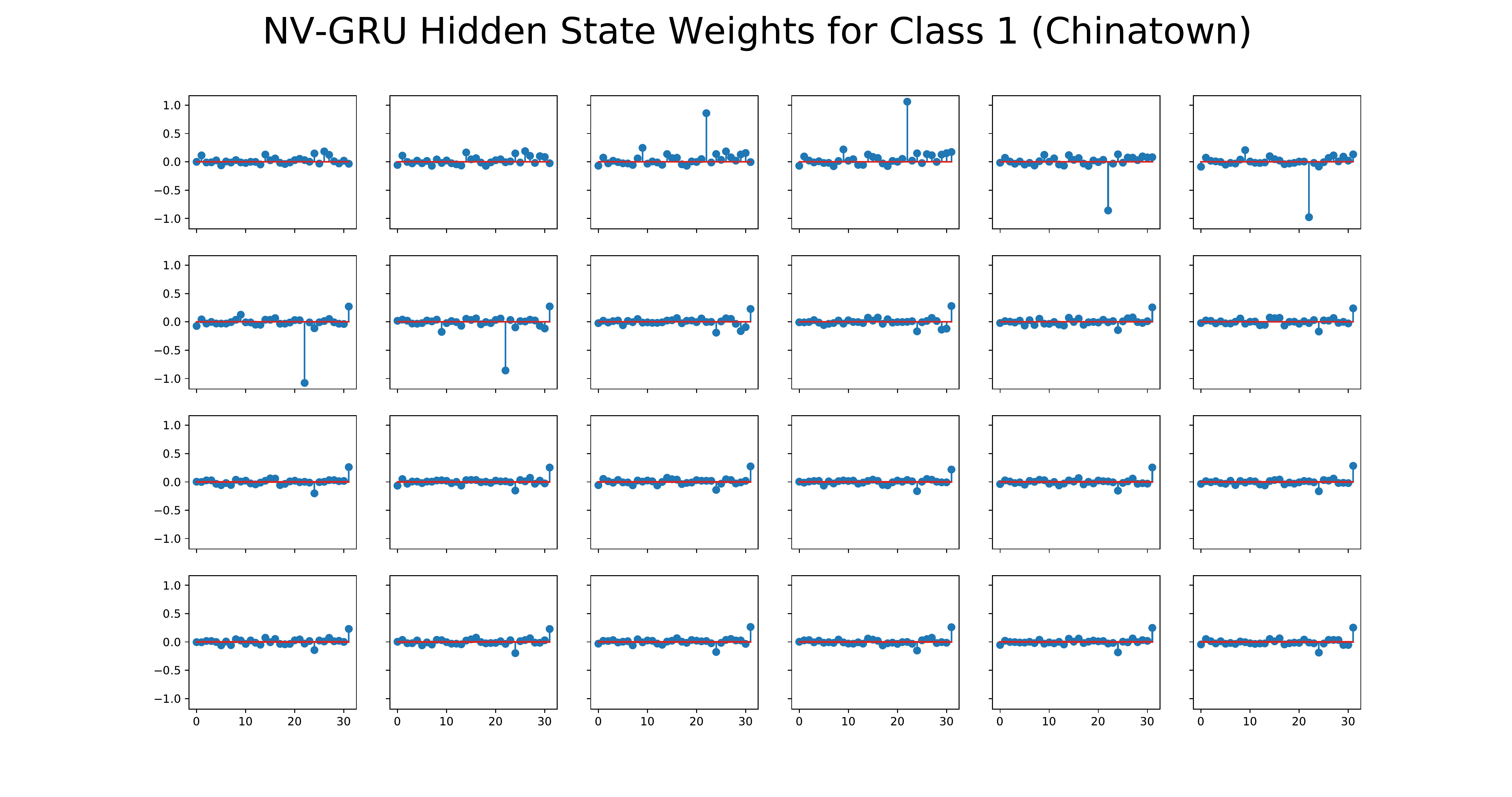}\hfill
    \caption{(Left) NV-GRU hidden state weights for class 0. (Right) NV-GRU hidden state weights for class 1. For the Chinatown dataset, it has 24 time steps and the first hidden state starts at the top left and going from left to right where the last hidden state is in the bottom right.}%
    \label{fig:chinatownNVHiddenStates}%
\end{figure}

One dataset from Table~\ref{table:1} is the Chinatown dataset that has 2 classes and the number of time steps is 24. Figure~\ref{fig:chinatownNV} shows the NV-GRU weights for both classes. The weights for each class are drastically different from each other. This notion makes sense because for binary classification, the objective would be to have the weights drastically differ. The positive weights for class 0 become the negative weights for class 1. Figure~\ref{fig:chinatownNVHiddenStates} shows the individual hidden state weights for more granular information. Since this dataset only has 24 time steps, it is easy to view the individual weights for every hidden state as opposed to other datasets. With the input size being one, there is not a lot of activity with the hidden states for every time step. It is interesting that for every time step there seems to be a gradual change when looking at the nearby hidden states at neighboring time steps.

Another dataset from Table~\ref{fig:fungiNV} is the Fungi dataset that has 18 classes and the number of time steps is 201. This dataset is very different from the Chinatown dataset, since we have multiple classes. Hence, one idea is to assess if there are weights from one class that is similar to another class and if there are weights from one class that is dissimilar to another class. To do this, we acquire the weights from each class and calculate the cosine similarity between all the classes. The cosine similarity is defined as $\frac{w_1^\top w_2} {||w_1||_2||w_2||_2}$ where $w_1$ is the weights of one class and $w_2$ is the weights of another class. In this scenario, class 5 is similar to class 16 while dissimilar to class 9. Figure~\ref{fig:fungiNV} shows the weights for each of these classes. Based on the cosine similarity, it is easier to perceive why class 5 and class 16 are similar. Plus, it is easy to identify that class 5 and class 9 are dissimilar in their weights. Hence, if there is a prediction that was meant for class 5 but went to class 9, this notion makes sense by looking at the weights.

We show different weights from different classes of the same dataset, but now we want to assess if NV-RNN, NV-GRU, and NV-LSTM prioritize on different time steps. This is done by using the Chlorine Concentration dataset, which has 3 classes and 166 time steps. From Table~\ref{table:1}, each of these NV models had different test accuracies. Figure~\ref{fig:chlorineNV} shows the different weights from these NV models for class 0. From Figure~\ref{fig:chlorineNV}, all of these NV models have a similar positive trend for the last few time steps. However, the middle time steps are where each of these NV models starts to differ in prioritizing certain time steps. Hence,  a benefit of the NV-RNN, NV-GRU, and NV-LSTM models is that you can interpret which time steps are critical based on the weight value. 

We use three datasets on how to interpret the NV-RNN models after they are trained. In addition, they can also explain their predictions since for each class, the weight values are provided and linked to all of the hidden states. Hence, with each NV-RNN model there is a formal manner to explain the predictions since we can acquire the hidden state values for each time step and the weights for each class.

\subsection{Case Studies}
We provide different case studies on how to use NV-RNN to inspect which time steps are prioritized within the application. With this insight, we observe how it provides additional understanding from the application.

\textbf{Different Weight Initializations}
RNNs can be difficult to train. From \cite{pascanu2013difficulty}, they state that early in the development of RNNs they would experience the vanishing gradient or exploding gradient problems. This led to the advent of GRUs and LSTMs. Even with the new gating architectures of GRUs and LSTMs, they can experience issues of learning. This discovery leads others to figure out how to improve the performance. Others such as \cite{le2015simple,talathi2015improving} found that initializing the $\mW$ weight matrix of the RNN can help in performance. Hence, we can use NV-RNN to inspect what is happening with the time step prioritization in regards to classification.

\begin{figure}[h!]

\centering
\includegraphics[width=.3\textwidth]{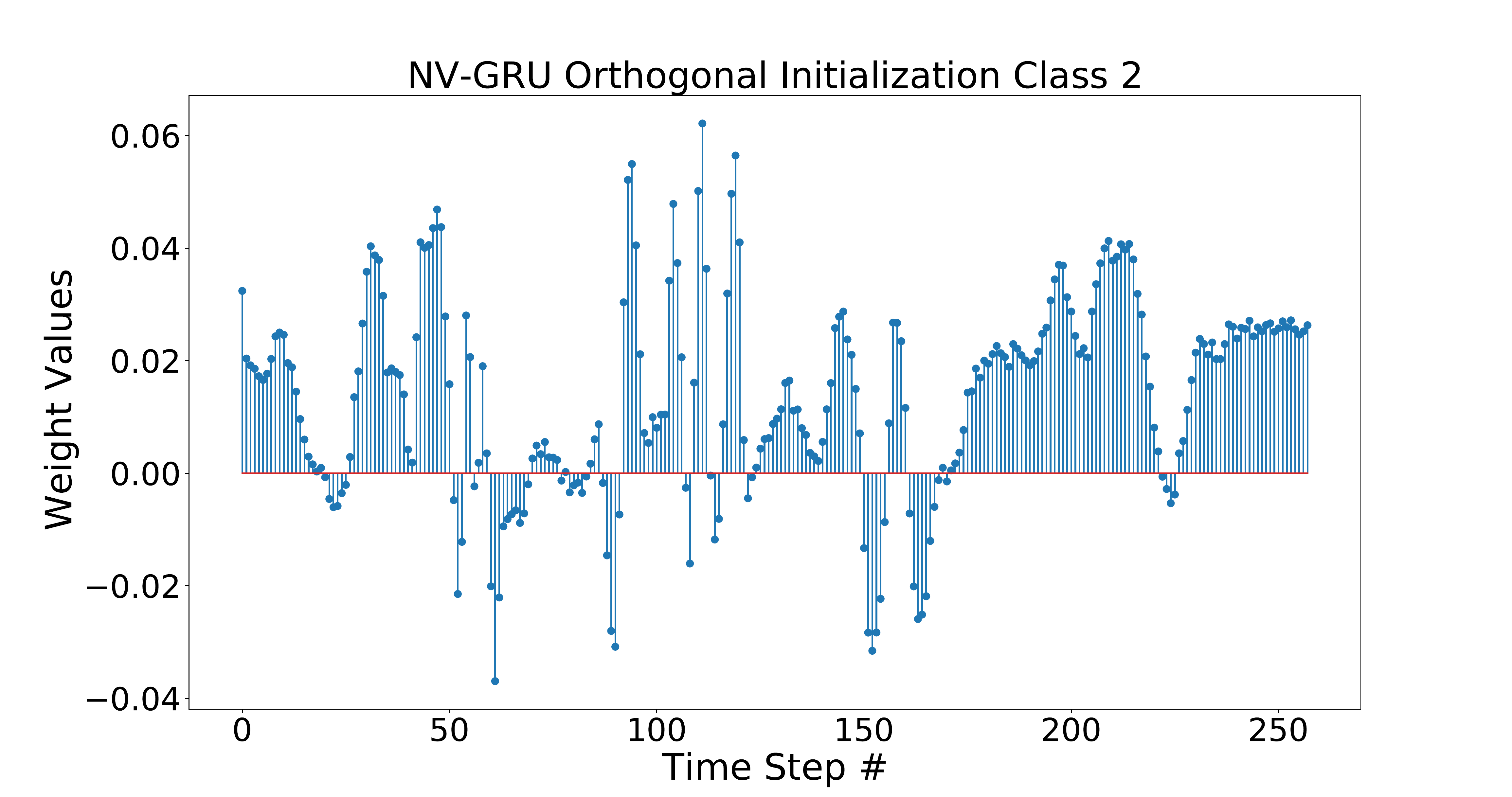}\hfill
\includegraphics[width=.3\textwidth]{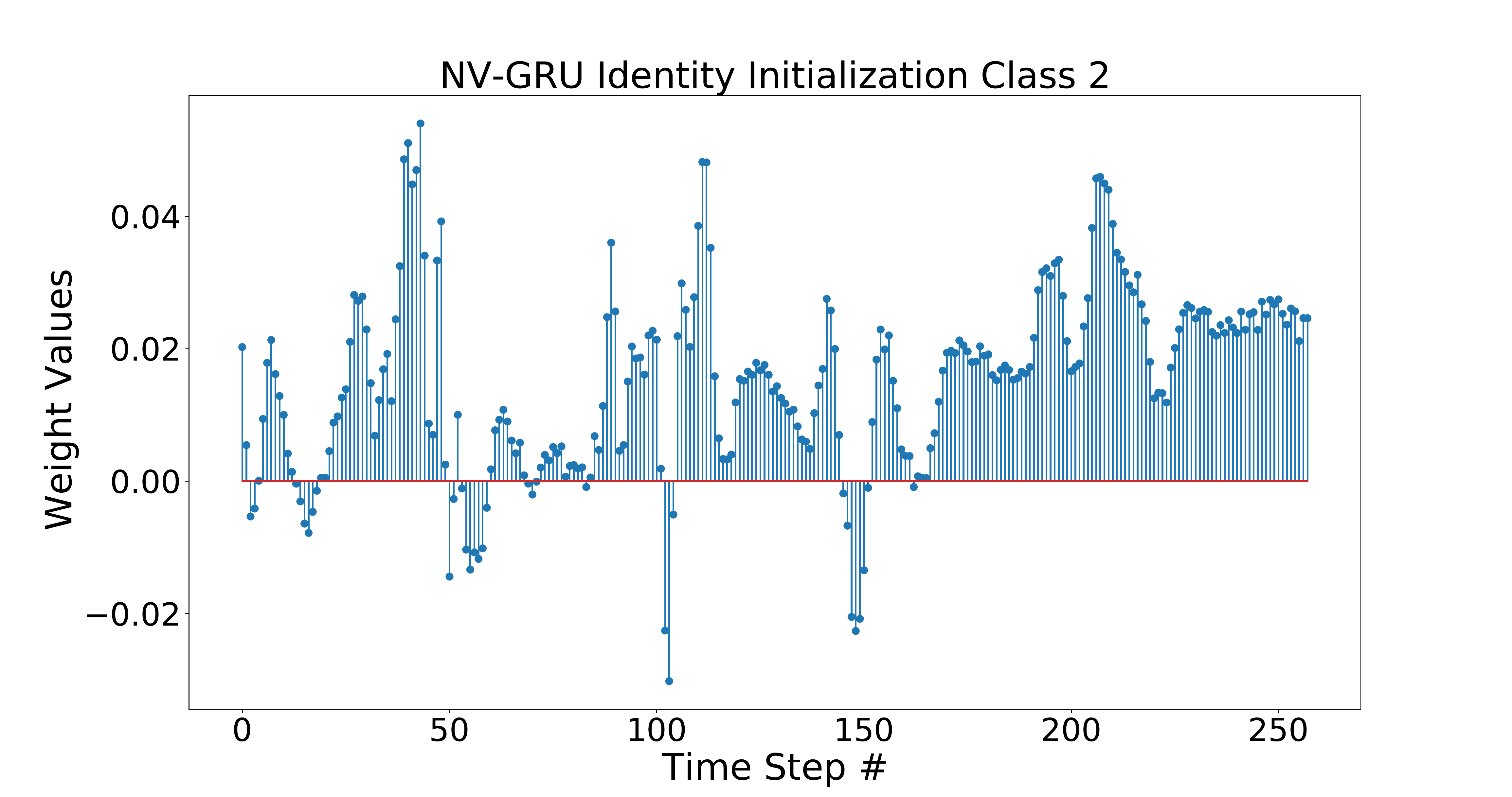}\hfill
\includegraphics[width=.3\textwidth]{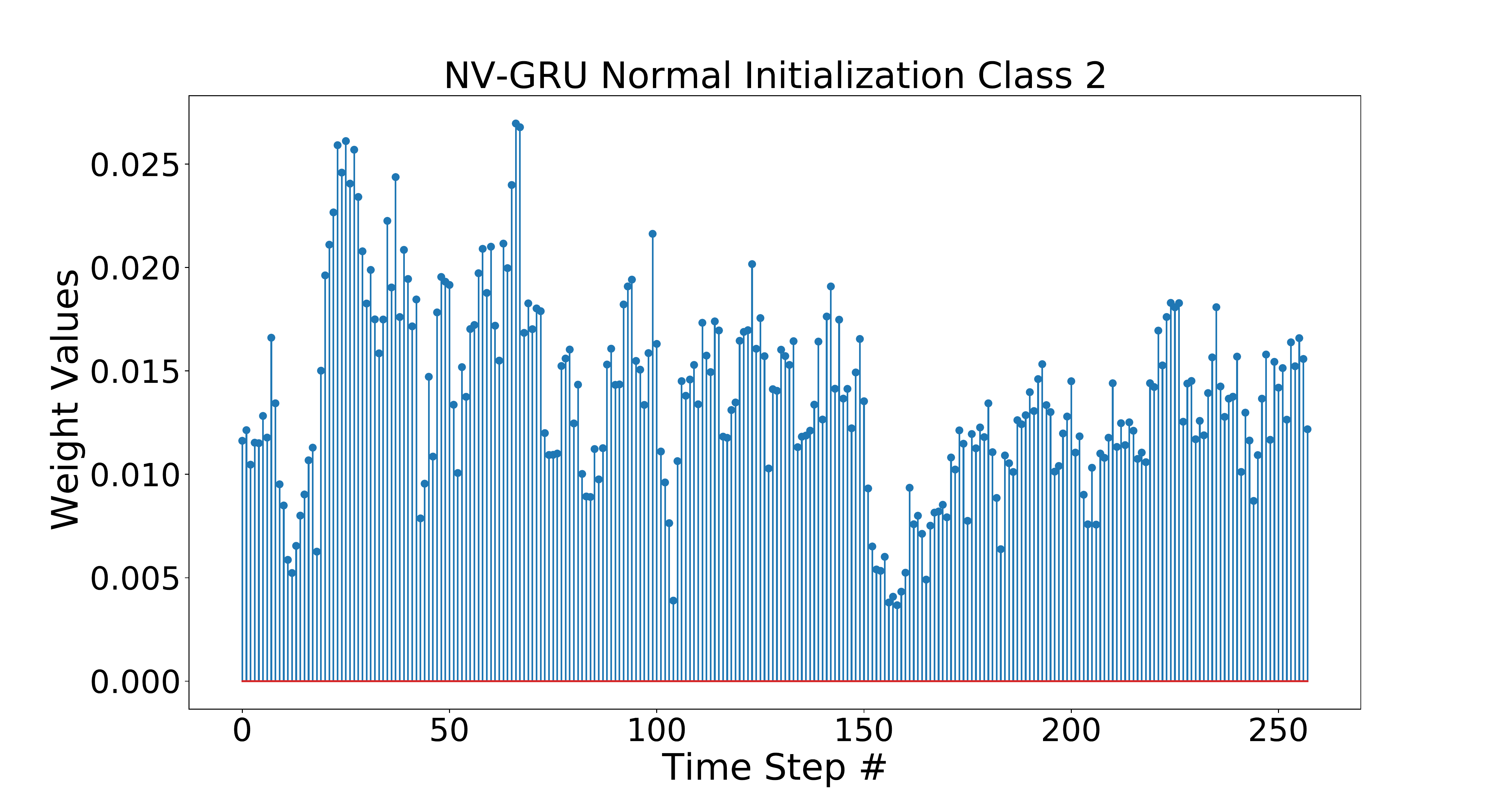}

\caption{(Left) Orthogonal. (Middle) Identity. (Right) Normal. With the same NV-GRU model but having 3 different weight initializations for $\mW$ of the NV-GRU network. With different weight initializations, the distribution of the weights mapping to the time steps is shown to be different. Hence weight initialization is important because even with similar performance, by looking inside of the linear weights can provide insight into what the network is deciding for classification in regards to class 2.}
\label{fig:weightsInit}

\end{figure}

On the InsectWingbeat dataset, we perform an experiment to vary the weight initialization on the hidden-to-hidden matrix. This matrix is different from the $\mV$ that is the linear classifier. We use the NV-GRU network to inspect how that can affect the decision-making process. The three different weight initializations are orthogonal, identity, and normal distributed. Figure~\ref{fig:weightsInit} shows the three different weight initialization schemes and you can notice that the weight initialization scheme using a normal distribution is focusing on different time steps compared to the same NV-GRU model but with different weight initializations.

\begin{figure}[h!]%
    \centering
\includegraphics[width=.3\textwidth]{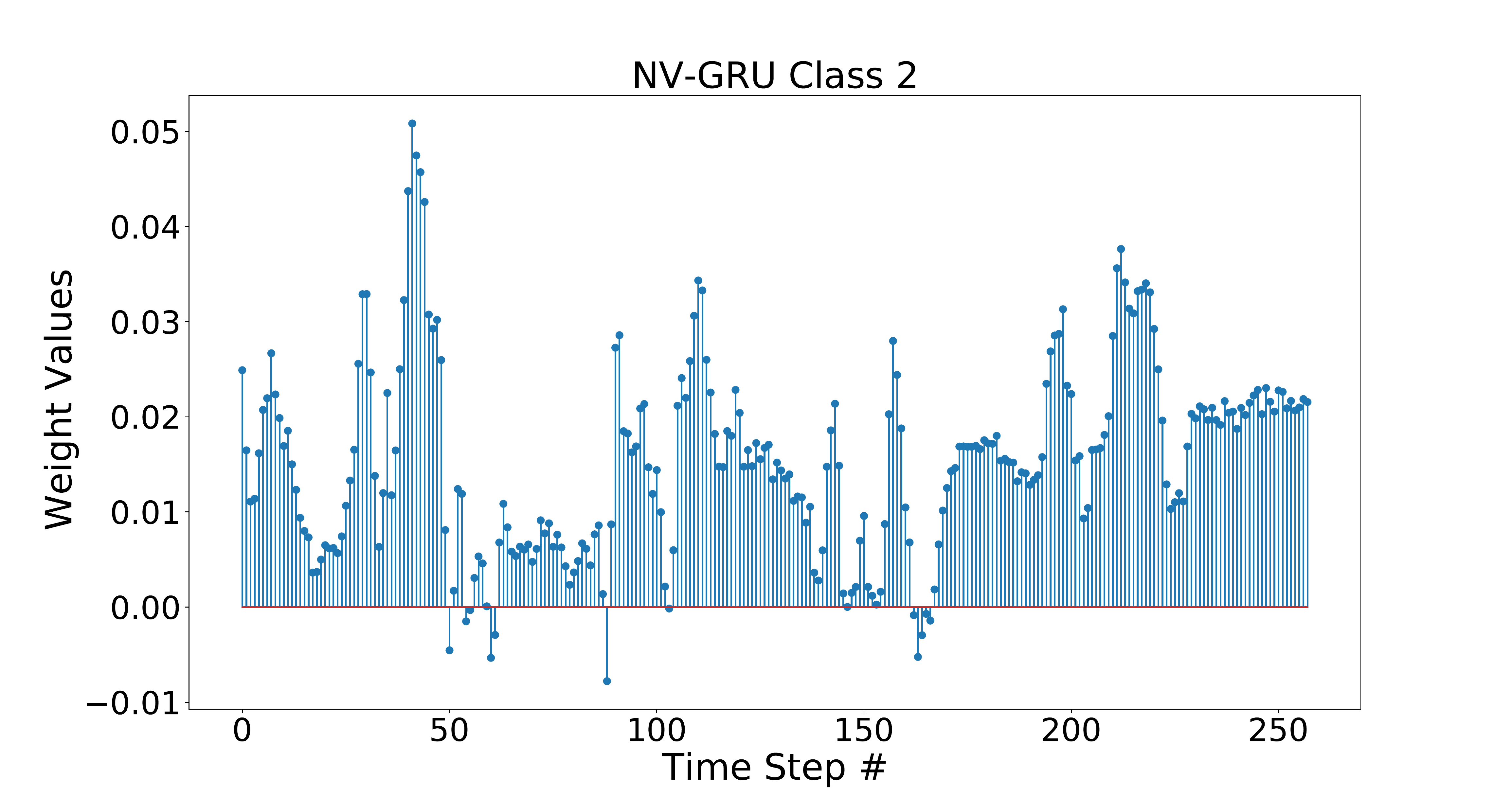}\hfill
\includegraphics[width=.3\textwidth]{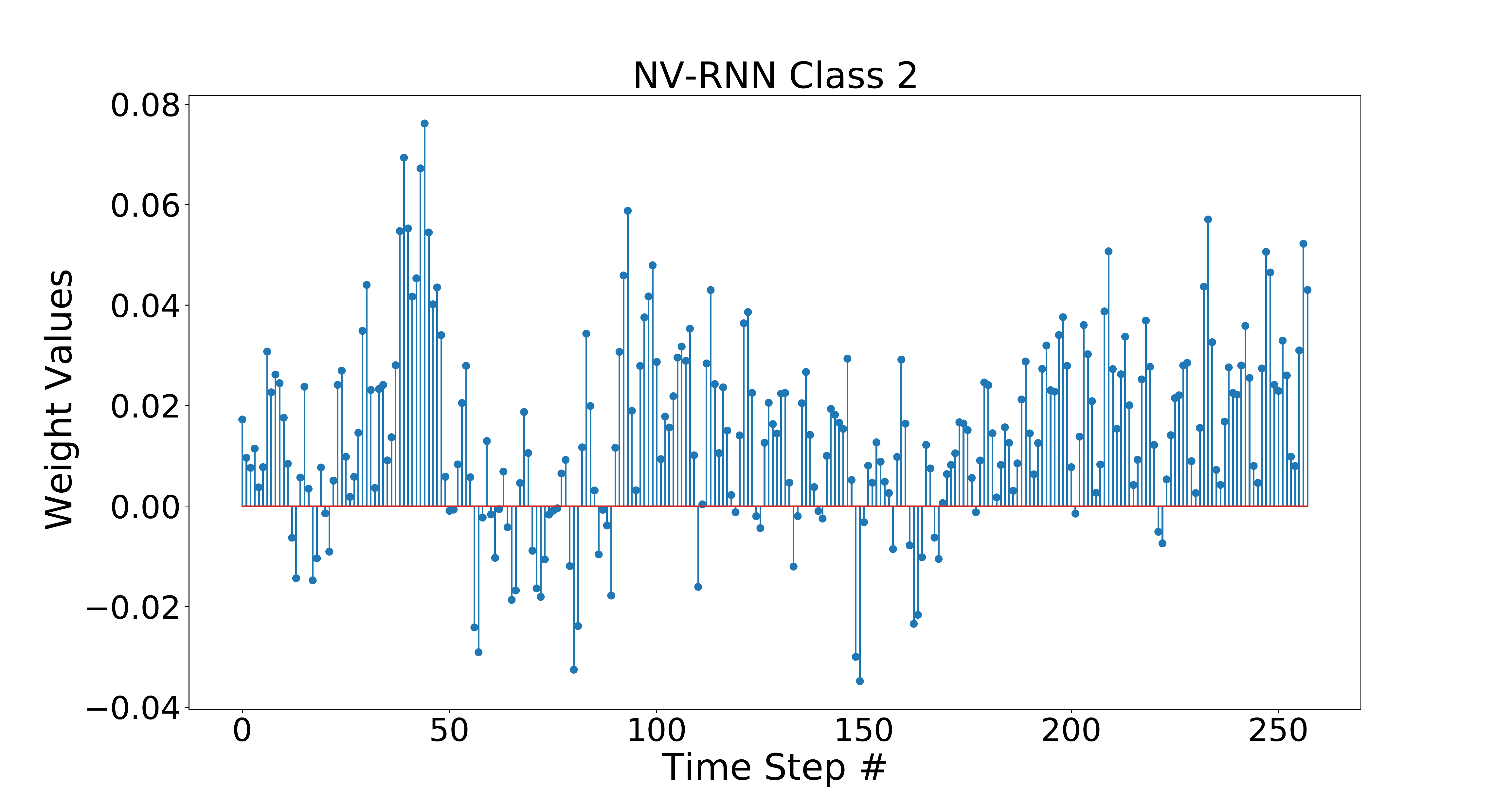}\hfill
\includegraphics[width=.3\textwidth]{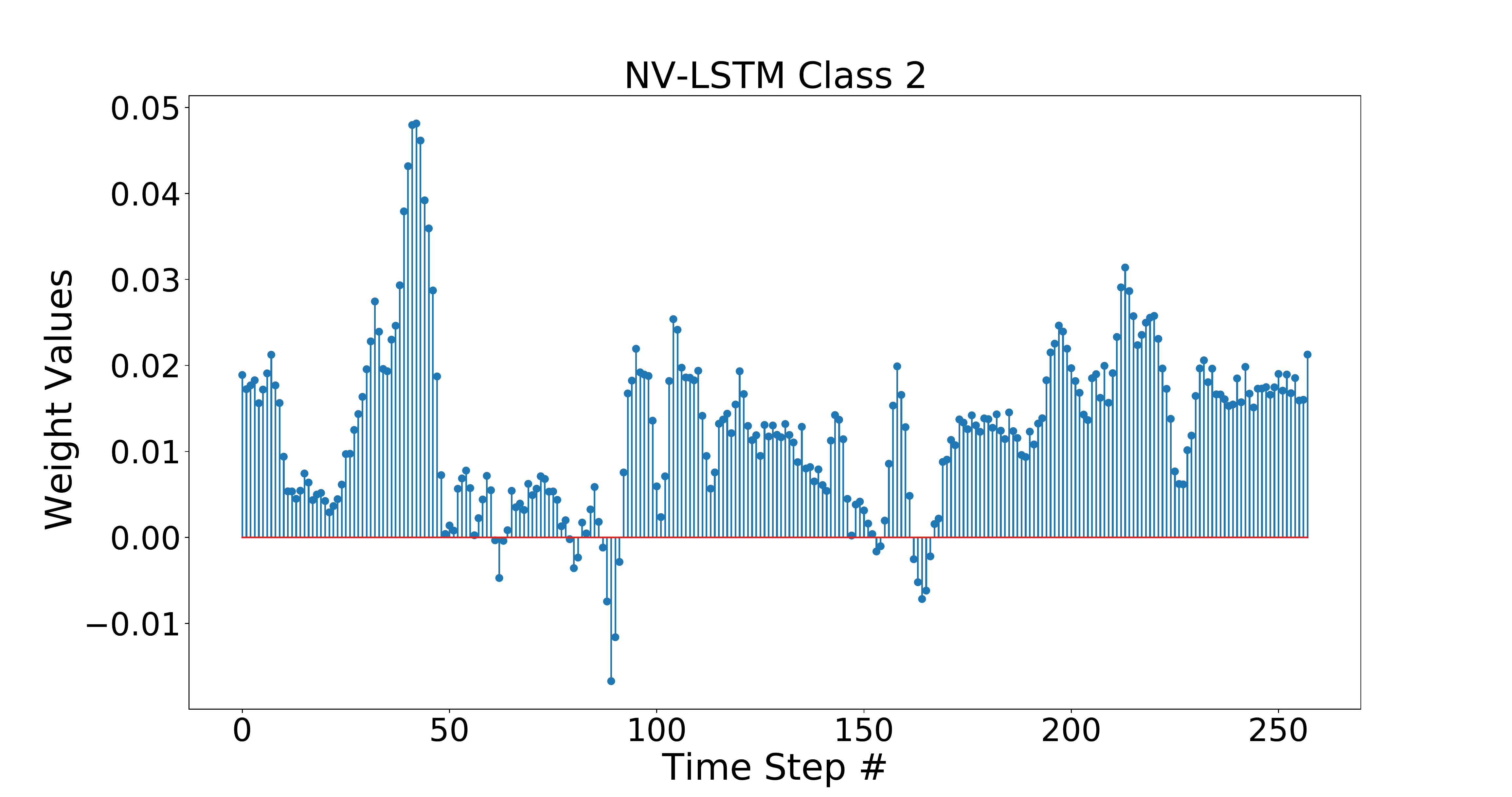}
    \caption{(Left) NV-GRU weights across time. (Middle) NV-RNN weights across time. (Right) NV-LSTM weights across time. All three different NV-RNN models display the weights of time to show for this class which time steps are the most significant. NV-GRU and NV-LSTM seem to behave in a similar manner while NV-RNN has a different way of prioritizing the time steps.}%
    \label{fig:insectNV}%
    
\end{figure}

The two weight initializations, orthogonal and identity, look very similar in terms of time step prioritization with Figure~\ref{fig:insectNV}. In Figure~\ref{fig:insectNV}, it is the default weight initialization for NV-RNN, NV-GRU, and NV-LSTM, which is uniform initialization. Thus it is interesting that even with two different weight initializations, those weight initializations look similar to each other. Yet, the normal initialization is vastly different in regard to the time step prioritization. Plus with the NV-RNN model, it can explain why the performance is bad by looking at the time step weight prioritization. Hence, even if the NV-RNN model is performing in a sub-optimal manner, we can inspect why it is performing in that manner. Using a traditional RNN, GRU, or LSTM cannot provide this information.  

\textbf{Increasing the Depth (More Layers)} 
\cite{pascanu2013construct} was one of the first works to take RNNs and make them deeper by stacking RNNs to increase the depth. In \cite{pascanu2013construct}, unfortunately in their work, they did not show any interpretation towards how each depth aids in classification. There was one work \cite{goh2017smiles2vec} that used a deeper GRU for a chemical task. They provide interpretability through a mask but unfortunately it lacks sufficient detail in how each RNN in the deep RNN is aiding in classification. This is where NV-RNN can aid and show how each RNN (layer) is prioritizing the time steps in terms of classification.

From Table~\ref{table:1}, we focused on one layer RNNs and NV-RNNs, but now we want to assess how increasing the depth of an NV-GRU model will show which time steps are prioritized. The question to inspect is if for every layer, will the time step prioritization be similar to the previous layers? Also, at a certain depth, will the time step prioritization be different from the previous layers? We can do this by looking at the weights of a NV-RNN model. In this analysis, we used the Rock dataset which has 2844 time steps and 4 classes. We have 4 different NV-GRU models where each one has varying depth. The first one starts at a depth of one and each additional model increases by an additional depth. Hence the last NV-GRU model has a depth of 4. 

When comparing the test accuracy of the different depths of NV-GRU to GRU and average pooling GRU, the NV-GRU models would outperform the GRU and average pooling GRU. When testing the performance among different hidden states like 32, 64, and 128, NV-GRU would have a test performance of \textbf{76\%}, \textbf{76\%}, and \textbf{76\%} for the depths of 2,3,4. This is the same performance as noted for a one-layer NV-GRU for the same dataset. For the GRU and average pooling GRU, the best test performance for depths of 2, 3, and 4 were 68\%, 72\%, and 66\%. Thus, for different depths, NV-GRU is outperforming.

\begin{figure}[h!]%
\centering
\includegraphics[width=.45\textwidth]{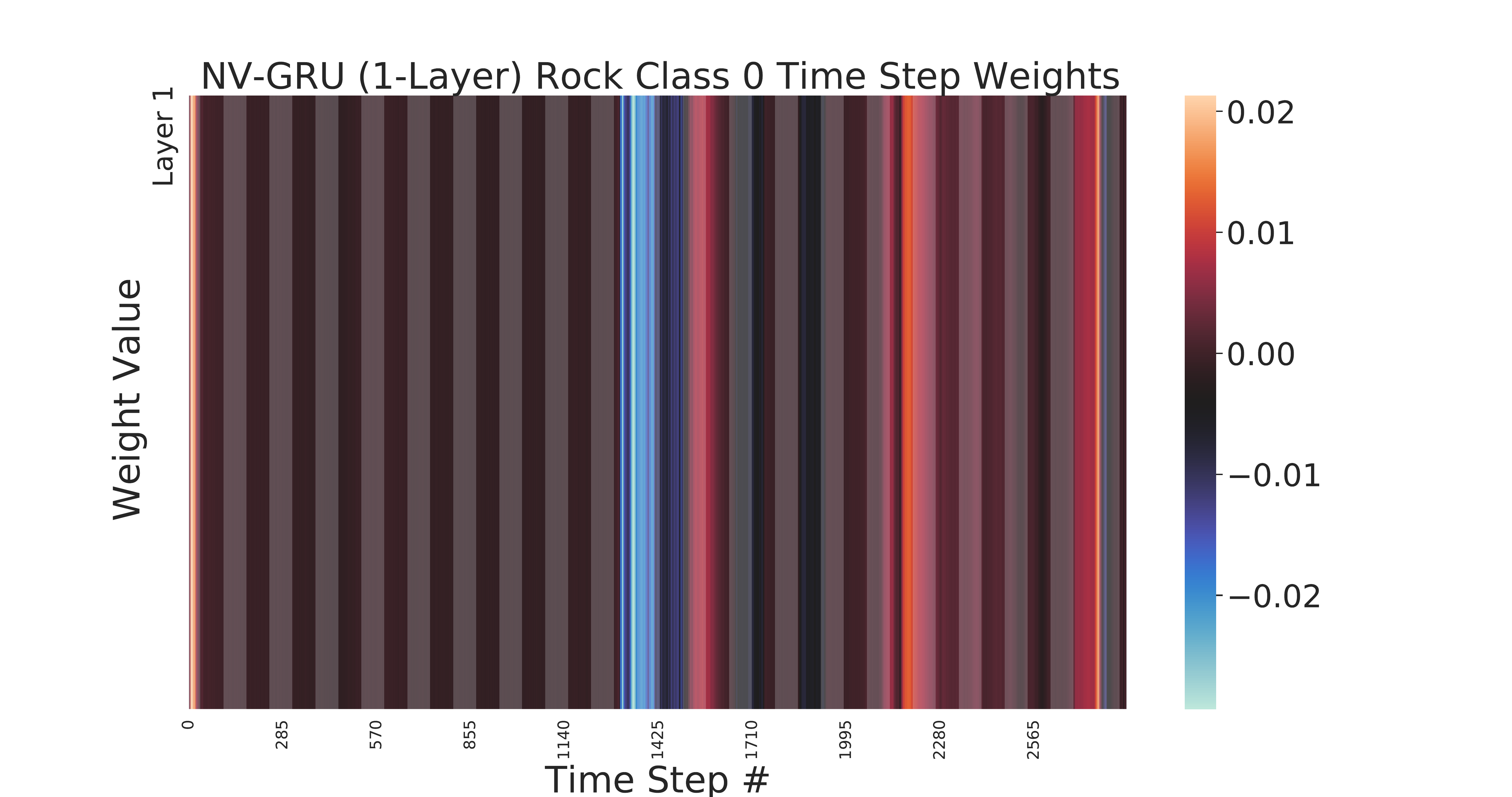}\hfill
\includegraphics[width=.45\textwidth]{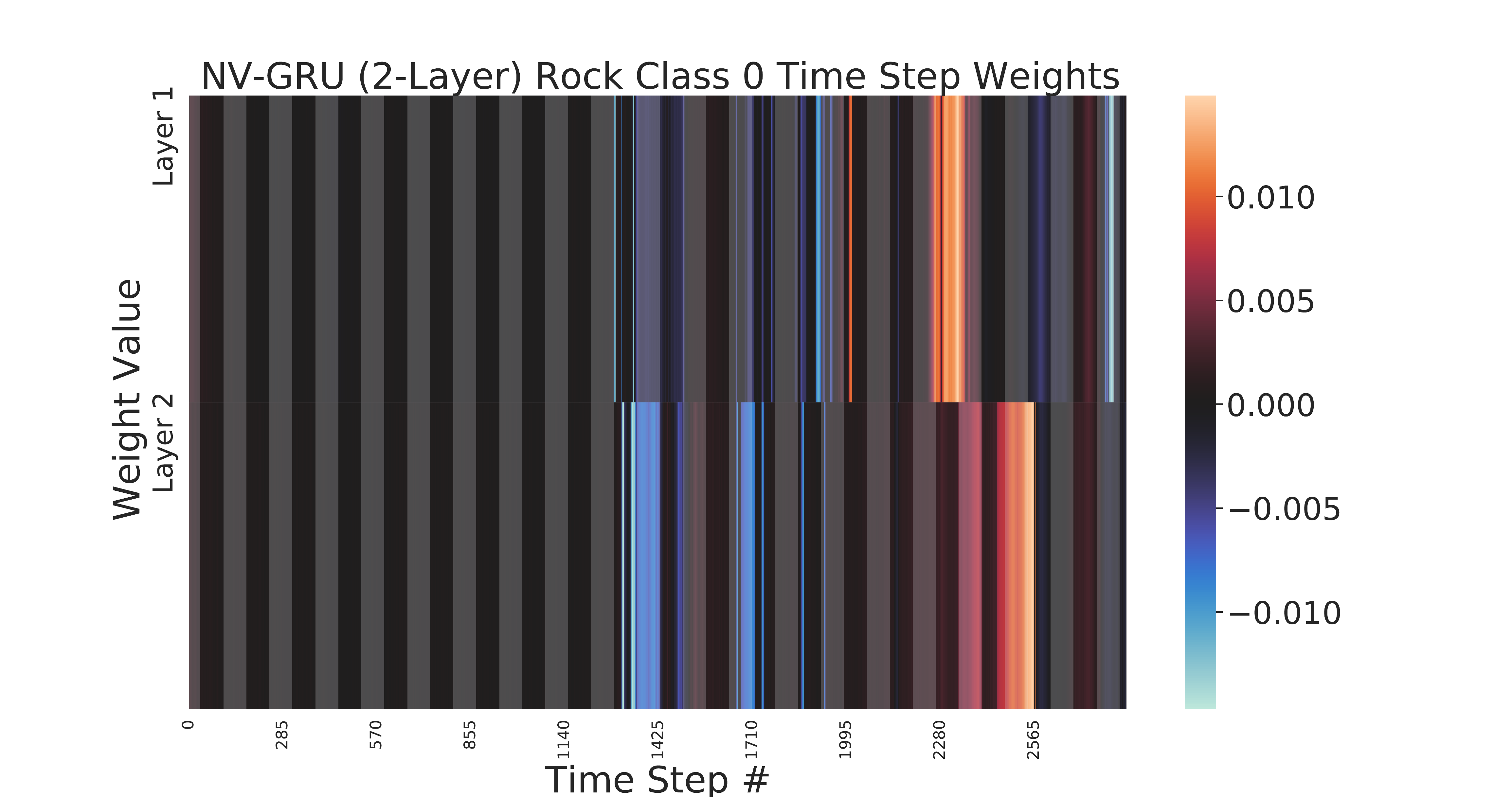}\hfill
\includegraphics[width=.45\textwidth]{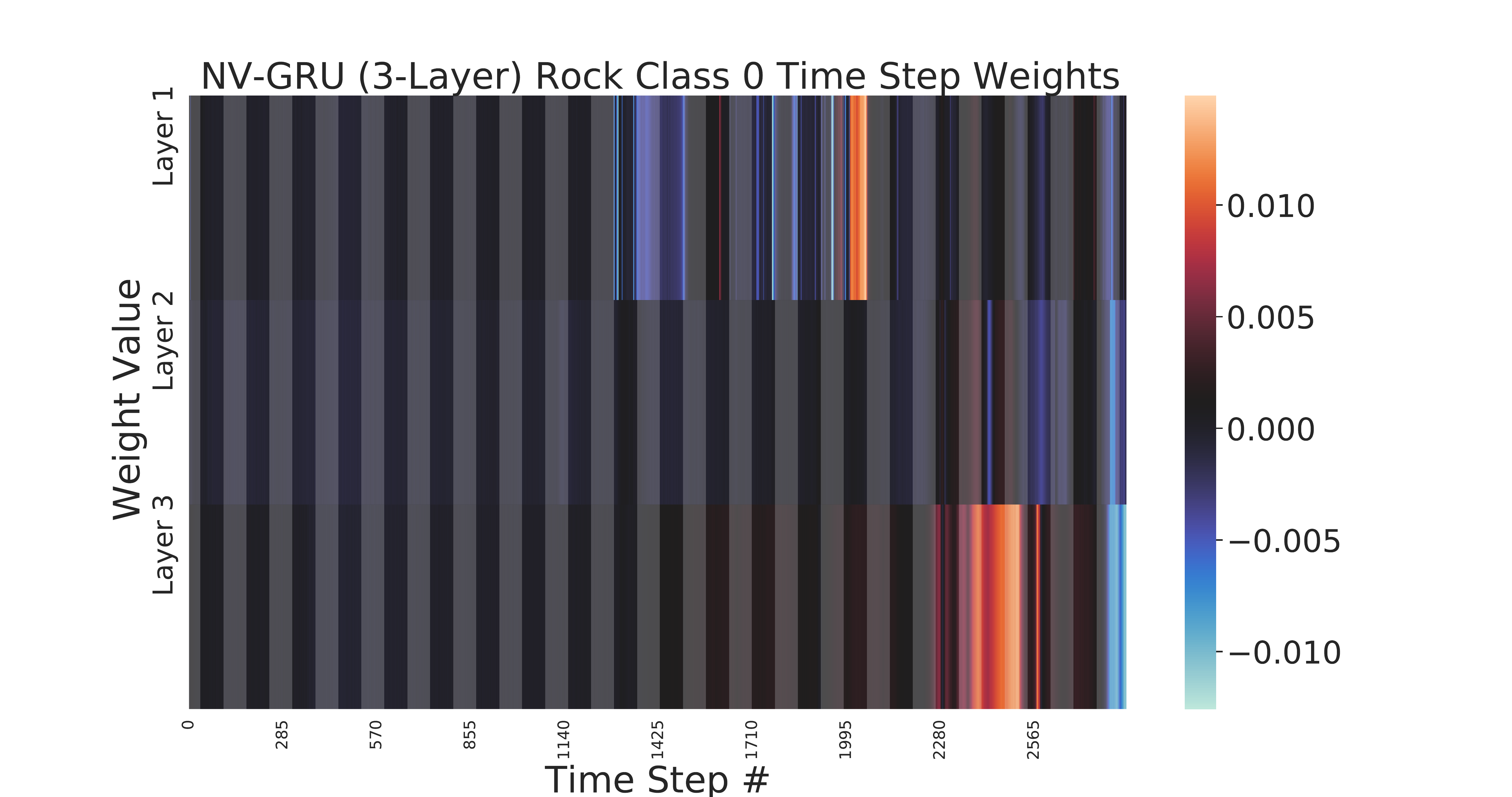}\hfill
\includegraphics[width=.45\textwidth]{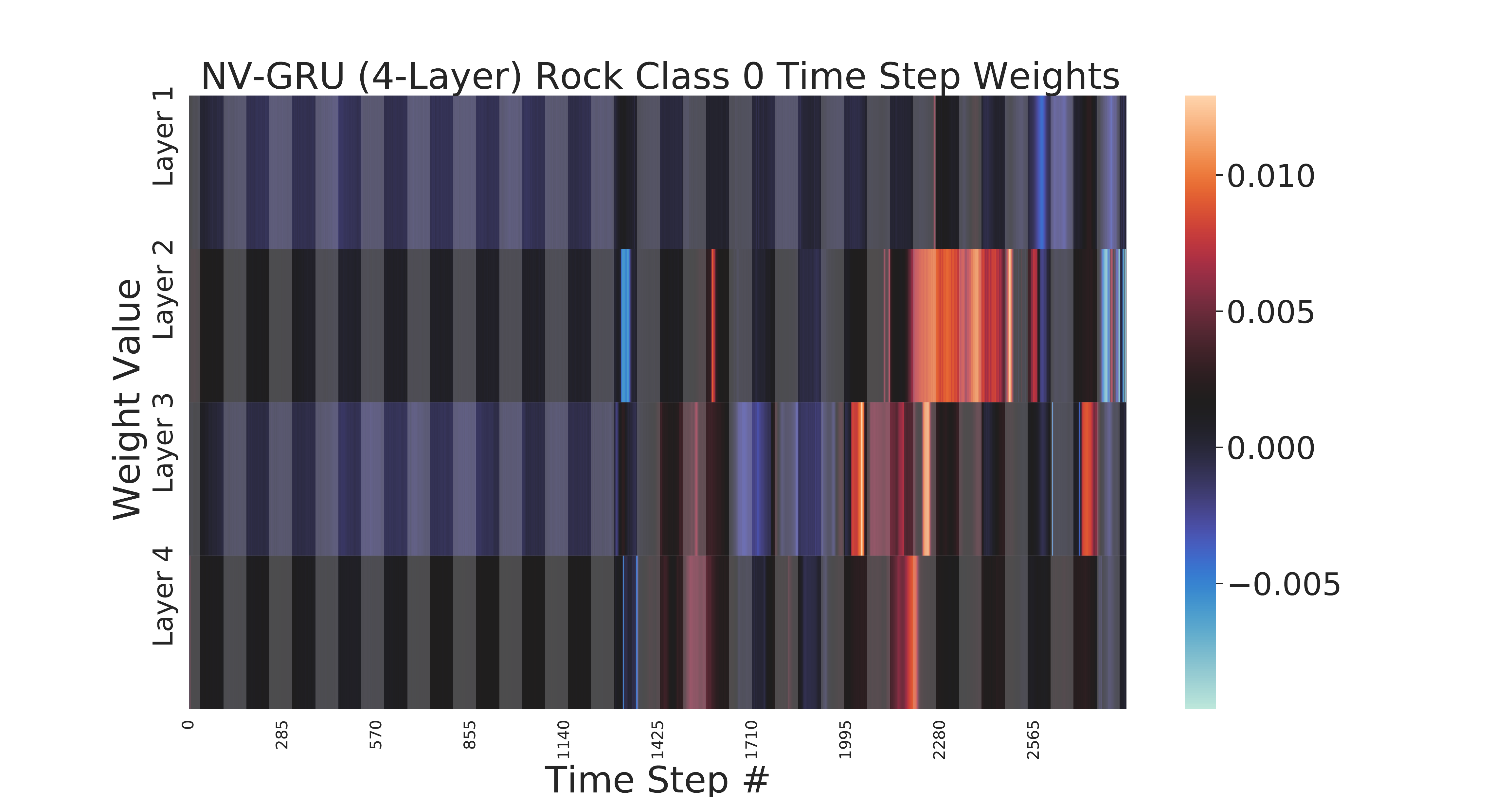}\hfill
    \caption{(Top Left) NV-GRU weights for class 0. (Top Right) NV-GRU (2 Layers) weights for class 0. (Bottom Left) NV-GRU (3 Layers) weights for class 0. (Bottom Right) NV-GRU (4 Layers) weights for class 0. By increasing the number of layers for the NV-GRU model, the prioritization of time steps is not consistently the same among all the layers. Each vertical column is the weight value at that time step.}%
    \label{fig:rockDepthNV}%
\end{figure}

Figure~\ref{fig:rockDepthNV} shows the time step weights for four different NV-GRU models with different depths. With NV-GRU, we can answer that question in a quantitative manner if the time step prioritization is similar among the different layers of the NV-GRU model. The answer is no and this makes sense since the input to the next layer is the previous layer's representation. For some time steps, there is some manner of consistency like with the NV-GRU models of 2, 3, or 4 layers where the beginning time steps are close to zero. Then the time step prioritization will vary towards the end of the time steps.

\textbf{Bidirectional RNNs} There are some RNNs that were developed that are bidirectional such that the RNN will learn in two directions. One direction is in a causal manner (forward) going from the beginning of the input to the end of it. While the other direction is in a non-causal manner (reverse) where it begins at the end of the input and ends at the beginning of the input. Equation~\ref{eq:bidirectional} details how an RNN will use both the forward and reverse hidden states. Thus the aspect for NV-RNN is to adapt it to the bidirectional variant and to inspect the weights. By inspecting the weights, we can assess how the time steps are prioritized in either direction. One question to answer is if the time steps of either direction will be symmetrical to each other?

There are works from \cite{ma2017dipole, schwab2017beat} that have used bidirectional RNNs to aid performance in their respective task. For interpretability, they use the attention weights to show how a given input is being classified. Yet in their interpretability, they cannot provide how each direction in the RNN is aiding for the classification. This lack of directional interpretability is where NV-RNN will help and can answer the questions mentioned above.

For the bidirectional analysis, we use an NV-GRU model that allows the bidirectional nature. We use the Rock dataset which has 2844 time steps and 4 classes.

When comparing a bidirectional NV-GRU with a bidirectional GRU and average pooling GRU, the bidirectional NV-GRU outperformed in test accuracy. Among the different hidden state sizes of 32, 64, and 128, the best test accuracy for bidirectional NV-GRU is \textbf{76\%} while for GRU is 72\% and average pooling GRU is 66\%. Even for the bidirectional variant, NV-GRU is outperforming.

\begin{figure}[h!]%
\centering
\includegraphics[width=.45\textwidth]{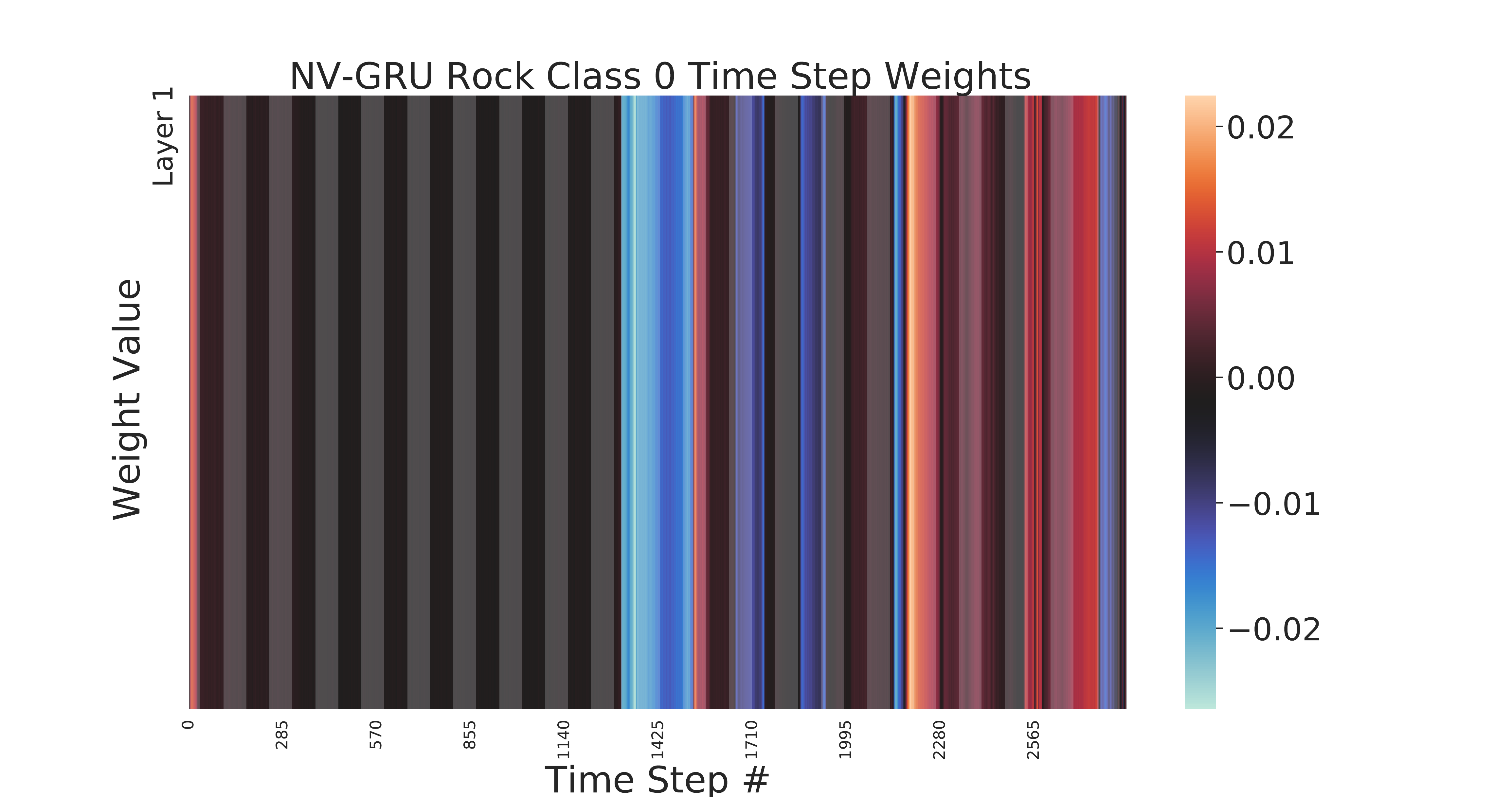}\hfill
\includegraphics[width=.45\textwidth]{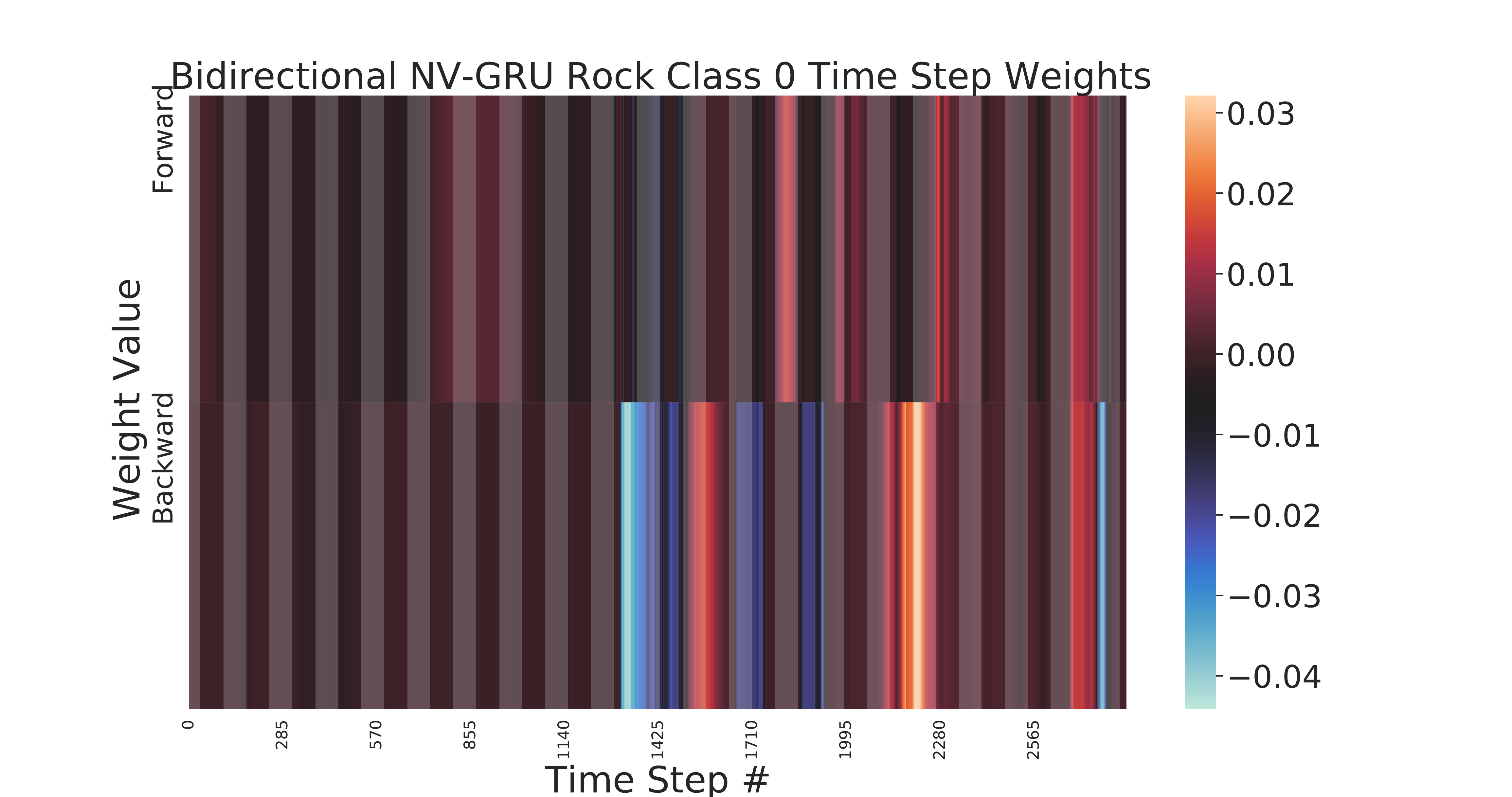}\hfill
    \caption{(Left) NV-GRU weights for class 0. (Right) Bidirectional NV-GRU weights for class 0. The forward and reverse prioritizations of time steps looks very similar. Each vertical column is the weight value at that time step.}%
    \label{fig:rockBiDirNV}%
\end{figure}

Figure~\ref{fig:rockBiDirNV} shows the time step prioritizations for the bidirectional NV-GRU. With this NV-GRU model, we can assess if the time steps prioritized from both directions will be symmetrical. The answer is that there is evidence that we do not see this notion and we can see that in a visual and quantitative manner (Figure~\ref{fig:rockBiDirNV}). Looking at the vertical columns for each direction, each direction does not have the same prioritization.

\textbf{Sentiment Analysis} 
The previous datasets that were shown consisted of time-series data. Now we will use a movie review dataset \cite{maas-EtAl:2011:ACL-HLT2011} with three embedding techniques, Word2Vec \cite{mikolov2013efficient}, Fasttext \cite{bojanowski2017enriching}, and GloVe \cite{pennington2014glove}, to convert each word into an embedding vector. Since NV-RNN is versatile, we can inspect how the weights for each class are prioritized for text data. In this dataset, the task is sentiment classification where one class is positive sentiment and the second class is negative sentiment.

Other works have provided different forms of understanding the differences of these embedding techniques. \cite{shi2014linking} show the similarities of GloVe and a skip-gram Word2Vec in a mathematical manner. \cite{nematzadeh2017evaluating} use a couple of evaluation tasks like correlation to show the differences or similarities of these embedding techniques. \cite{shin2018interpreting} uses eigenvector analysis to compare the embedding vectors to assess how the words cluster together. Yet, all of these techniques cannot link the time steps to the class, which is what NV-RNN will show.

Table~\ref{table:2} shows the performance and in this scenario, NV-GRU performs on par. Note that for this dataset, there are reviews of variable length so for a GRU the last hidden state per review is the input to the linear classifier. However, for NV-GRU, padding has to be applied since the input for the linear classifier has to be of fixed size. Hence, it does explain the small drop in performance. As for padding, we only used the first 1,000 words of the review. With reviews that are smaller than 1,000 words, zero padding would be applied. Hence, there can be a good parameter to pad the reviews. Even with this disadvantage, NV-GRU is still on par and only loses one to two percent of test accuracy.

\begin{figure}[h!]%
\centering
\includegraphics[width=.45\textwidth]{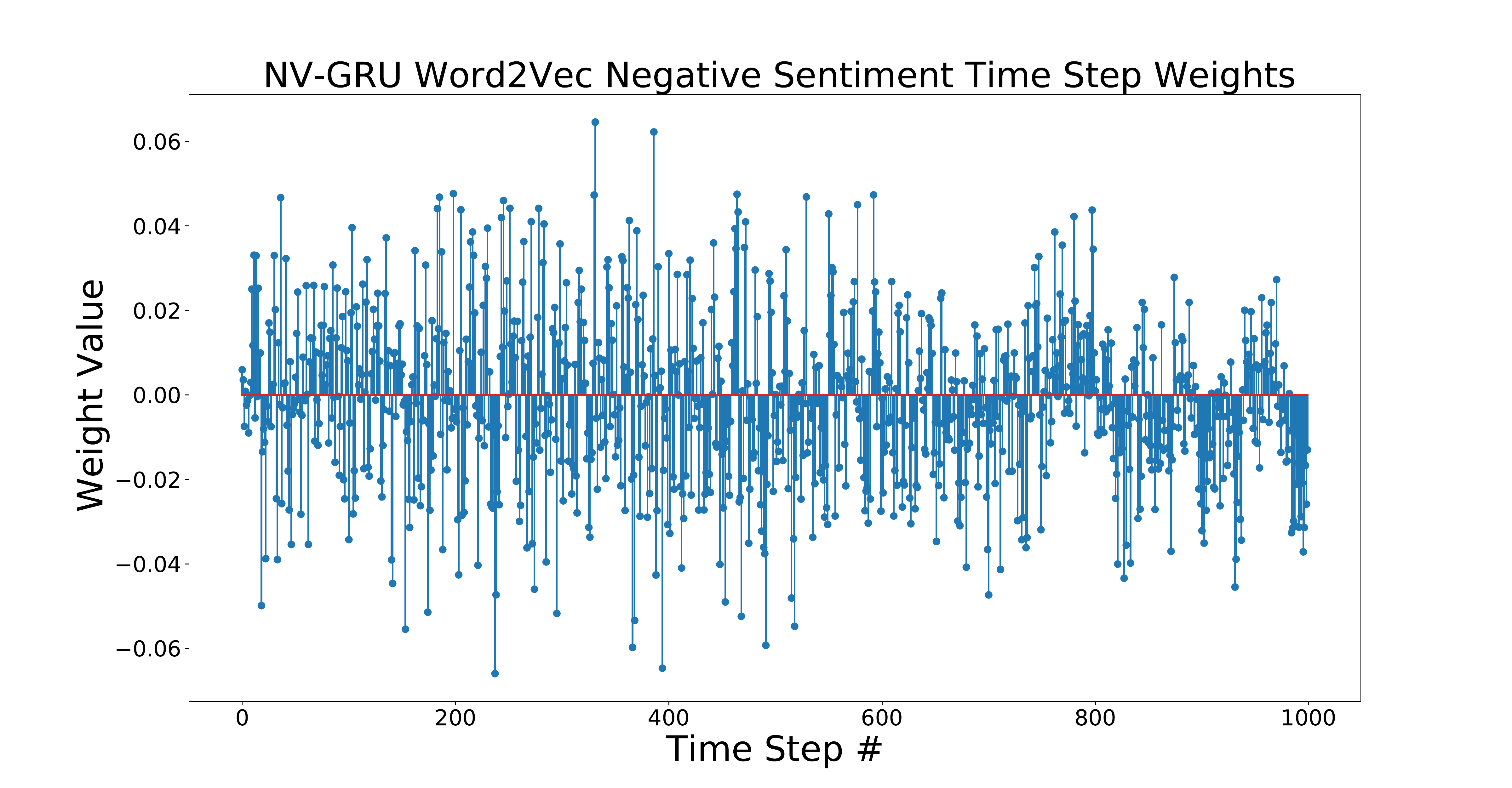}\hfill
\includegraphics[width=.45\textwidth]{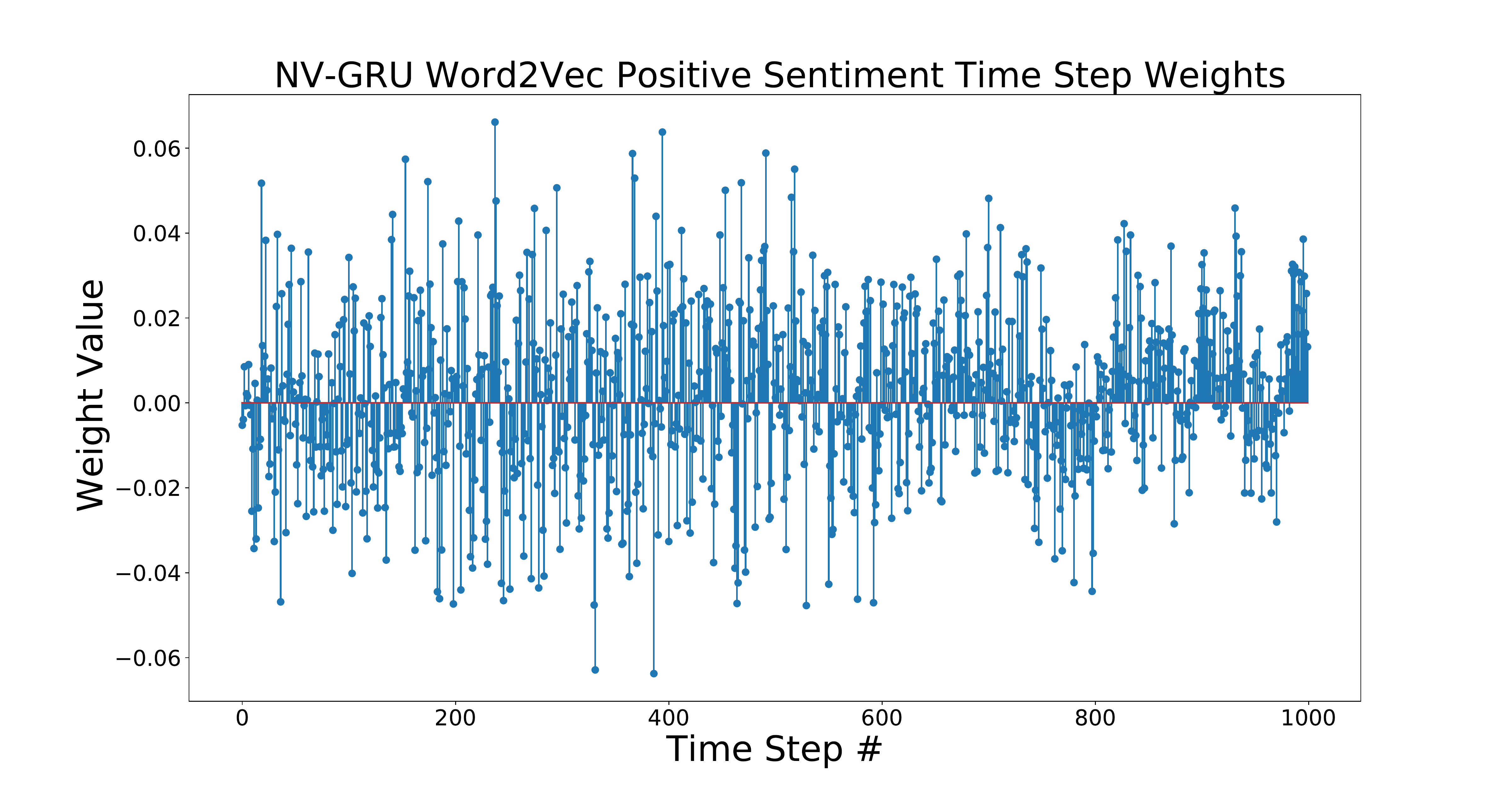}\hfill
    \caption{(Left) NV-GRU weights for class 0. (Right) NV-GRU weights for class 1. For the Movie Review dataset, there are two classes and the weights for each class are drastically different. In addition, the input is the Word2Vec embedding of each word per time step. Compared to continuous time-series data, the prioritization of time steps is not as smooth compared to datasets with continuous data.}%
    \label{fig:movieW2VNV}%
\end{figure}

\begin{figure}[h!]%
\centering
\includegraphics[width=.45\textwidth]{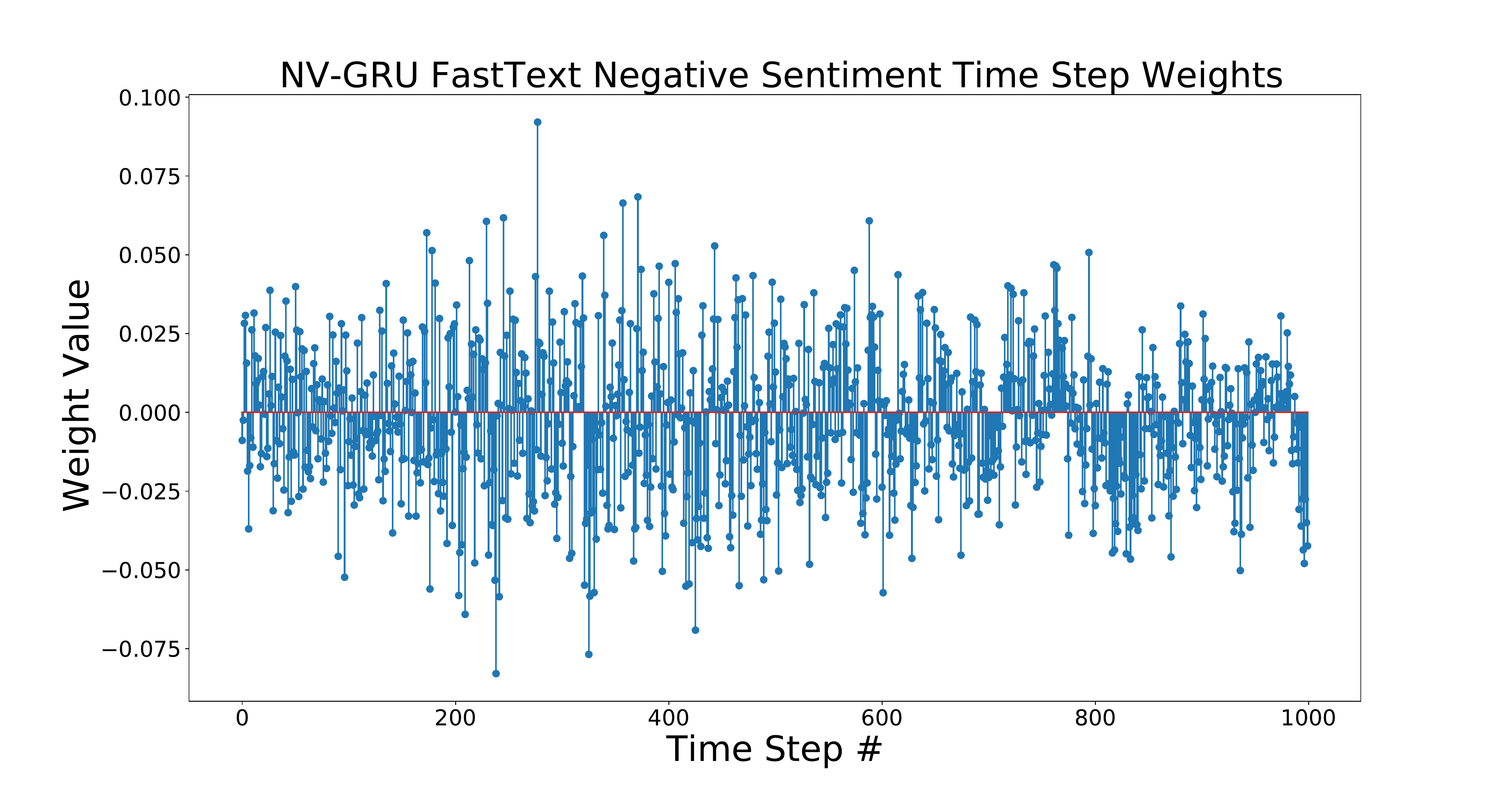}\hfill
\includegraphics[width=.45\textwidth]{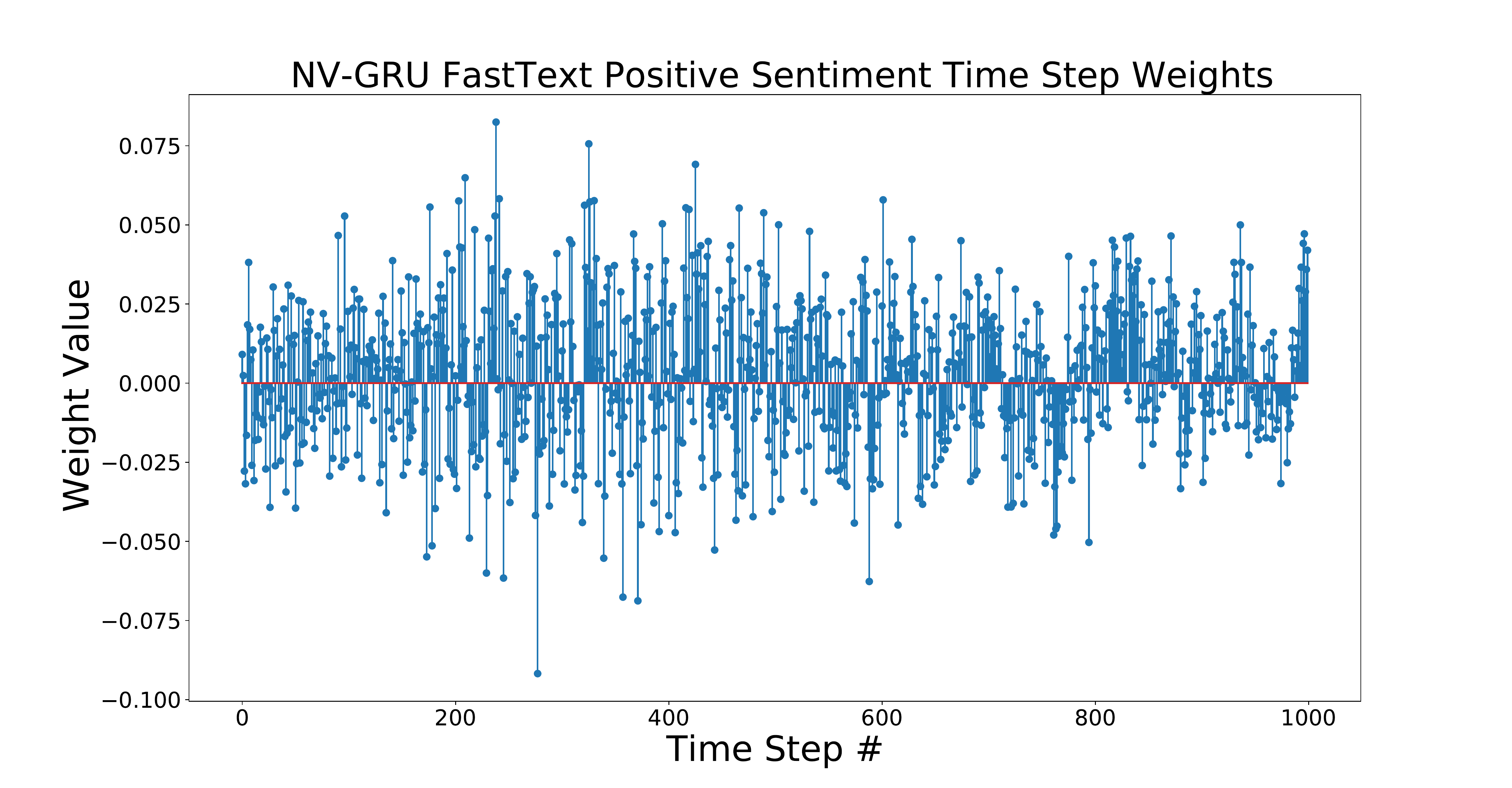}\hfill
    \caption{(Left) NV-GRU weights for class 0. (Right) NV-GRU weights for class 1. For the Movie Review dataset, there are two classes and the weights for each class are drastically different. In addition, the input is the FastText embedding of each word per time step. Compared to continuous time-series data, the prioritization of time steps is not as smooth compared to datasets with continuous data.}%
    \label{fig:movieFTNV}%
\end{figure}

\begin{figure}[h!]%
\centering
\includegraphics[width=.45\textwidth]{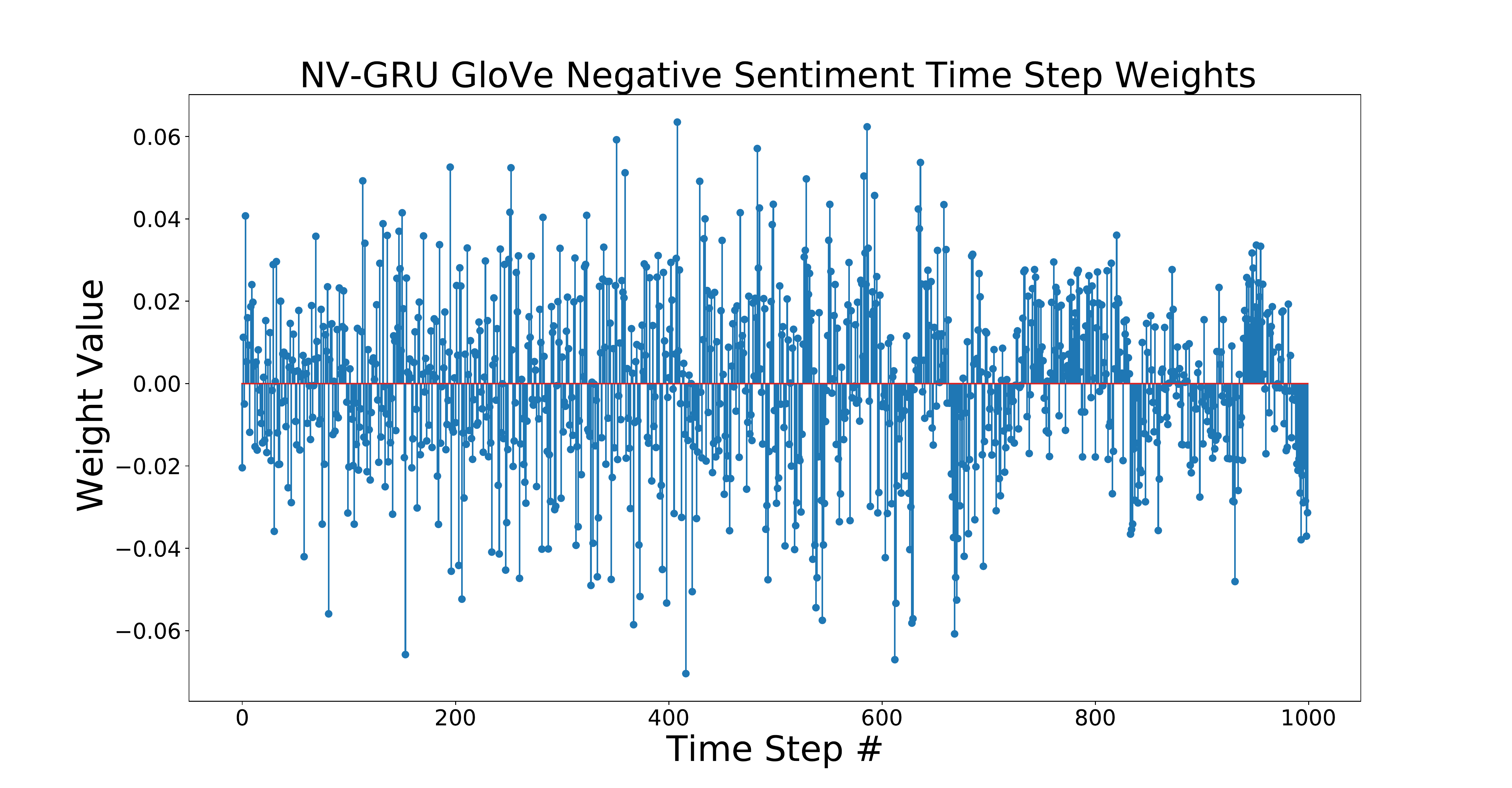}\hfill
\includegraphics[width=.45\textwidth]{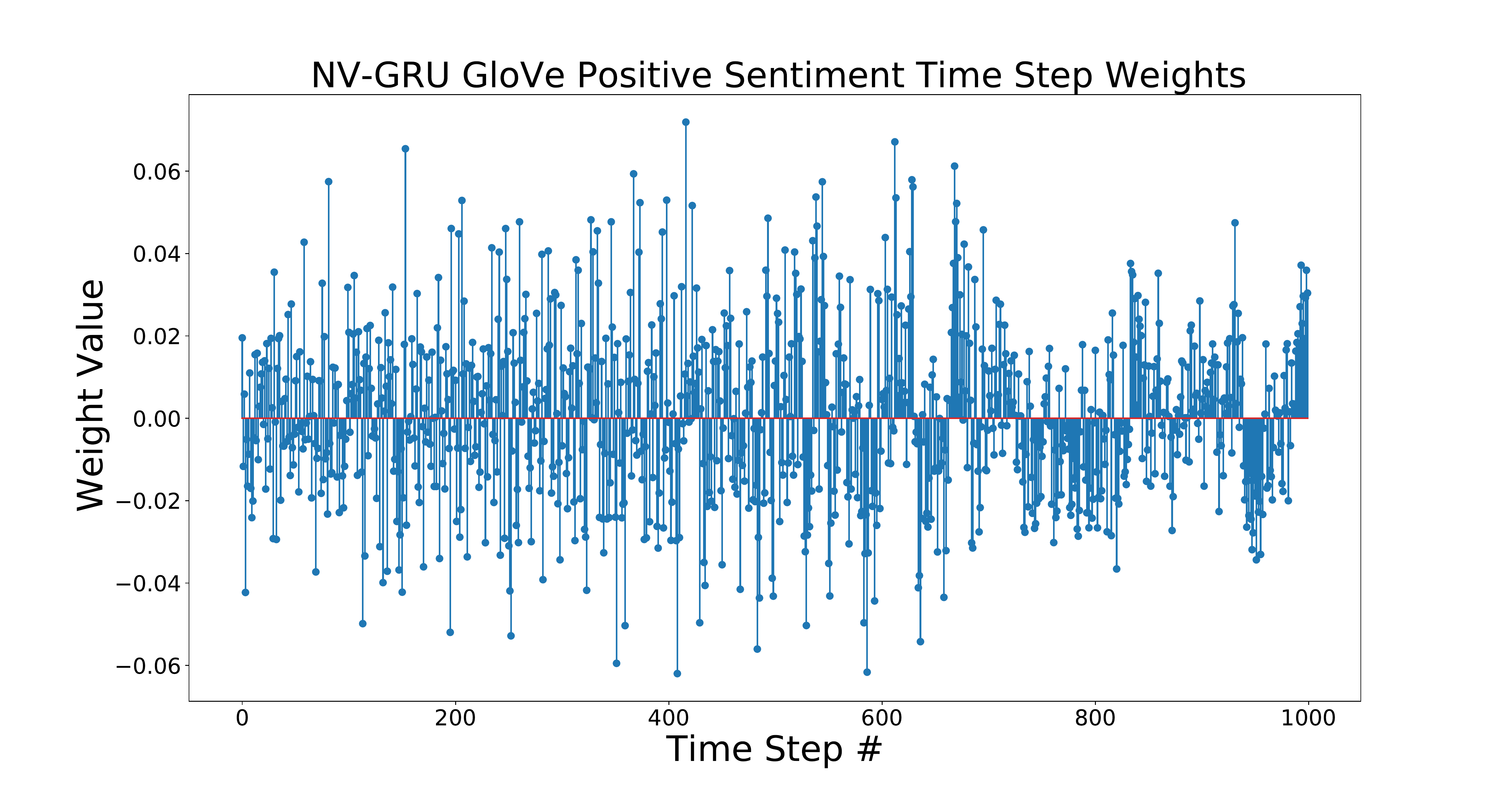}\hfill
    \caption{(Left) NV-GRU weights for class 0. (Right) NV-GRU weights for class 1. For the Movie Review dataset, there are two classes and the weights for each class are drastically different. In addition, the input is the GloVe embedding of each word per time step. Compared to continuous time-series data, the prioritization of time steps is not as smooth compared to datasets with continuous data.}%
    \label{fig:moviegloveNV}%
\end{figure}

Figures~\ref{fig:movieW2VNV}, \ref{fig:movieFTNV}, and \ref{fig:moviegloveNV} display the weights for each class. Even though there are three different embedding techniques, Word2Vec, FastText, and GloVe, it seems that the prioritization for the time steps in each embedding looks to be quite similar in the broad general sense. For the negative sentiment class, it is interesting how with all the embedding inputs, the last time steps are negative. This does make sense since FastText is a continuation of Word2Vec except for the case that FastText will approximate words that are not in the dictionary of words that it had learned. It is interesting how the GloVe embedding looks to have similar broad time step prioritization since it is a matrix factorization technique compared to Word2Vec and FastText. There are very specific differences based on the magnitudes but in a general manner all the embedding techniques for the input will result in the same general trend for each class in terms of the weights.

From Figures~\ref{fig:movieW2VNV} and \ref{fig:movieFTNV}, we provide another experiment to see which words are tied to the highest weights based on the time step index. To reduce the amount of reviews, we only look at the top 5 reviews for each class based on the pre-softmax score. In addition, we look at the top 10 weights per class. When inspecting the words for each class, we find pronouns and prepositions. However, one particular notion we inspect is particular words for each class. For the negative sentiment class, we observe the words, \textbf{bad}, \textbf{dodgy}, \textbf{unpleasant}, and \textbf{pointless} from the time steps with the highest weights. For the positive sentiment class, we observe the words, \textbf{mature}, \textbf{happy}, \textbf{perfect}, and \textbf{top}. We would hope to observe this notion because if the class is centered around negative sentiments, then the words that should be linked to the class would be bad words. The same notion applies to the positive sentiment class. Also, we did observe that for the positive sentiment class, we did not notice any negative sentiment words like bad or terrible.

\textbf{Video Action Recognition}
In this application, the input to the network is a collection of images from a video. The task is to predict which of the actions is presented in the video where the dataset is UCF11 \cite{liu2009recognizing}. In Table~\ref{table:3}, we show that the NV models are outperforming and we inspect the NV-CNN-GRU model. Note that in this scenario we are concatenating both the CNN filter units and the GRU hidden states.

There have been works such as \cite{dong2017improving,meng2019interpretable} to provide interpretability with video action recognition with CNN-LSTM models. Yet, one of the main issues is that they cannot explain which part, the CNN or LSTM is contributing to the classification. This is where NV-CNN-GRU is able to provide both interpretability and explainability.

\begin{figure}[h!]

\centering
\includegraphics[width=.4\textwidth]{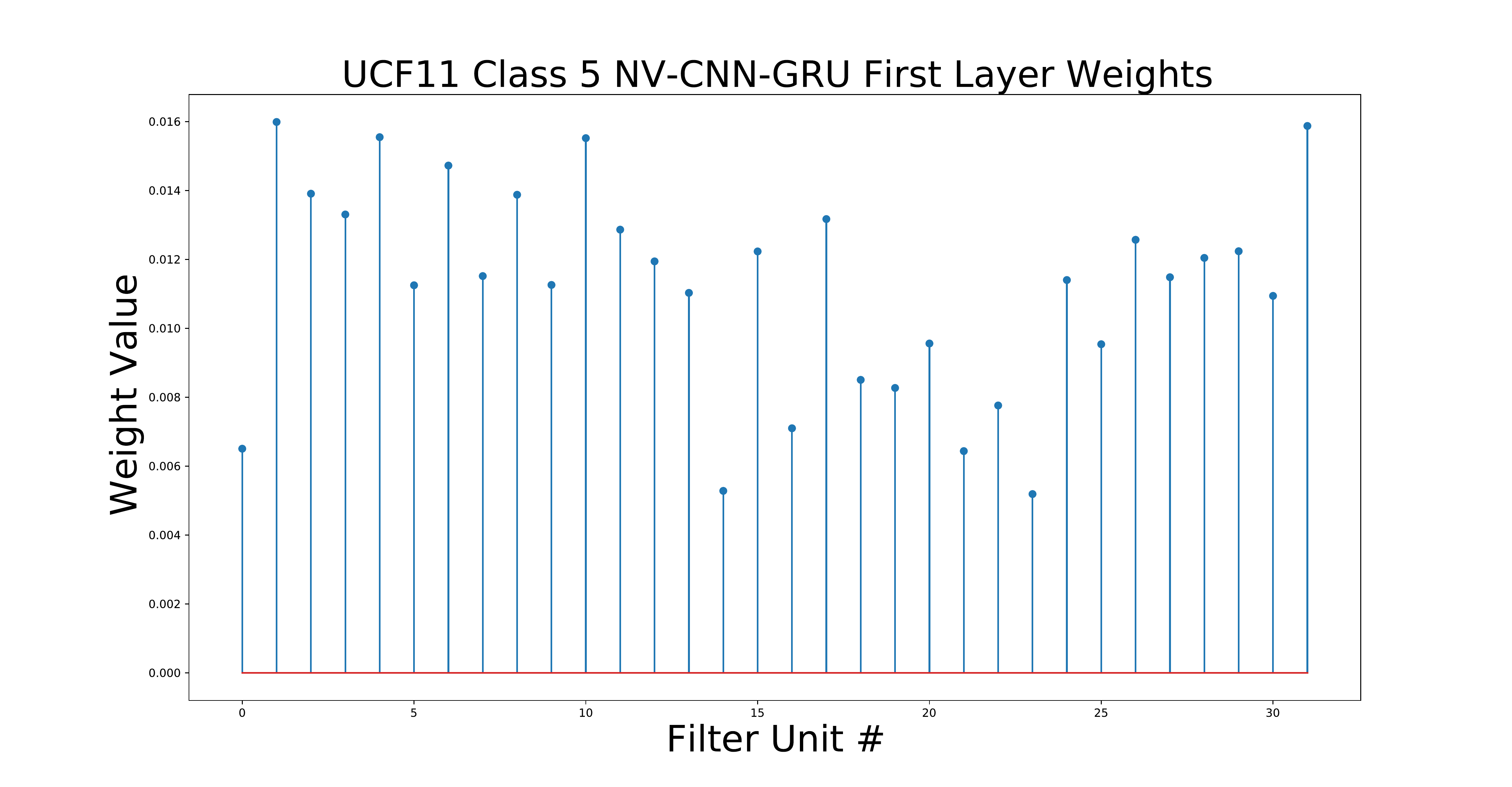}\hfill
\includegraphics[width=.4\textwidth]{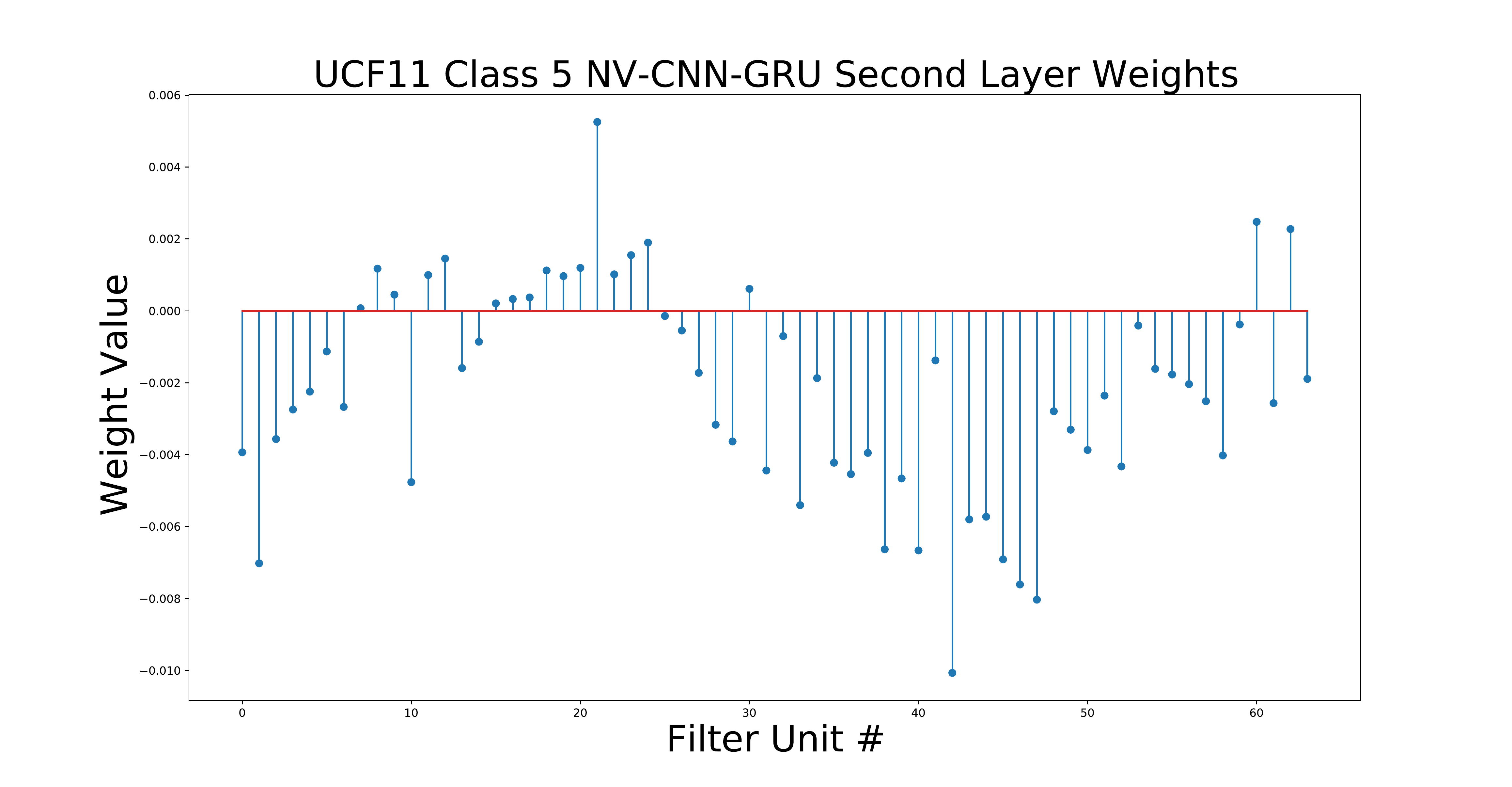}\hfill
\includegraphics[width=.4\textwidth]{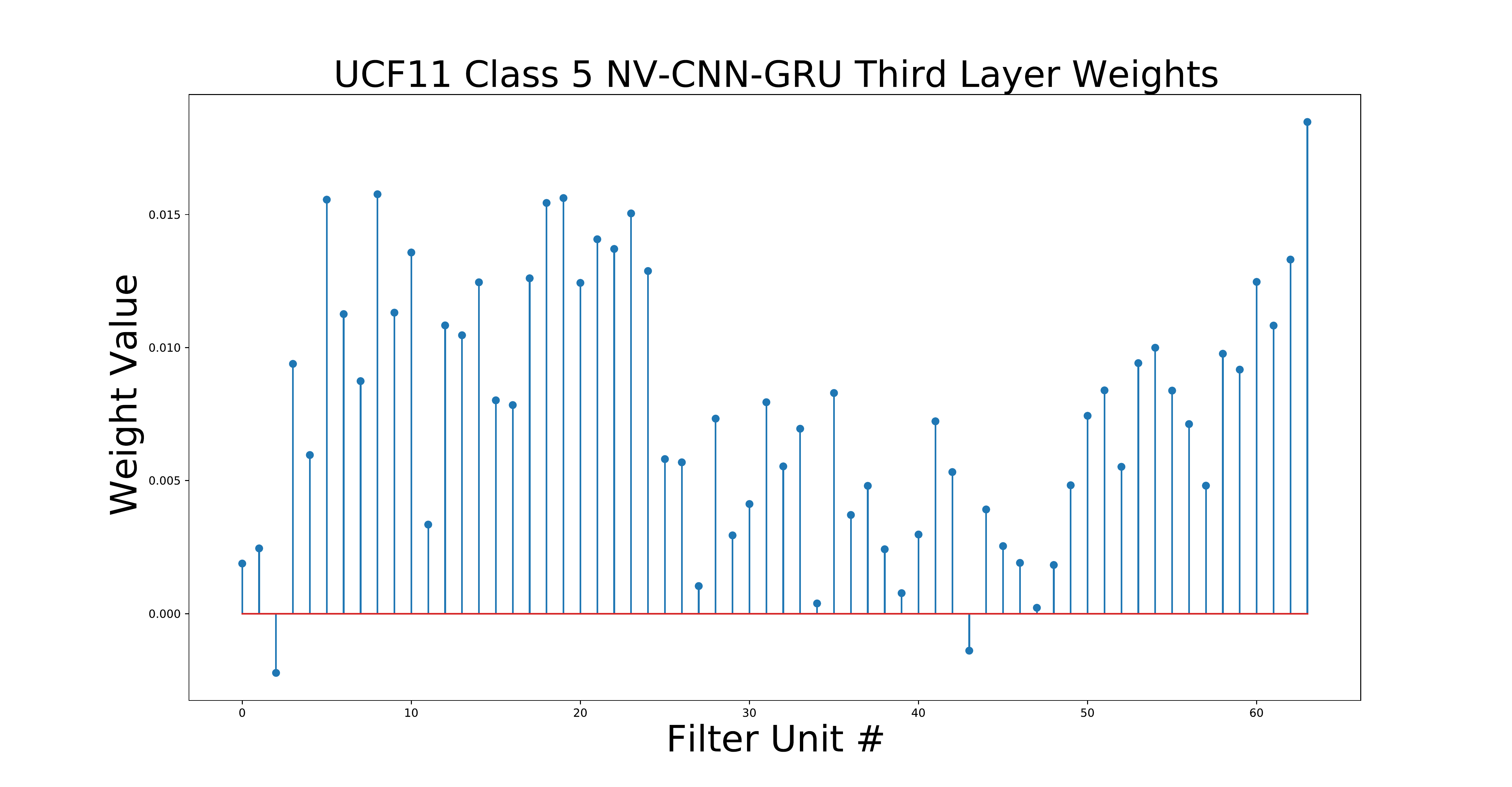} \hfill
\includegraphics[width=.4\textwidth]{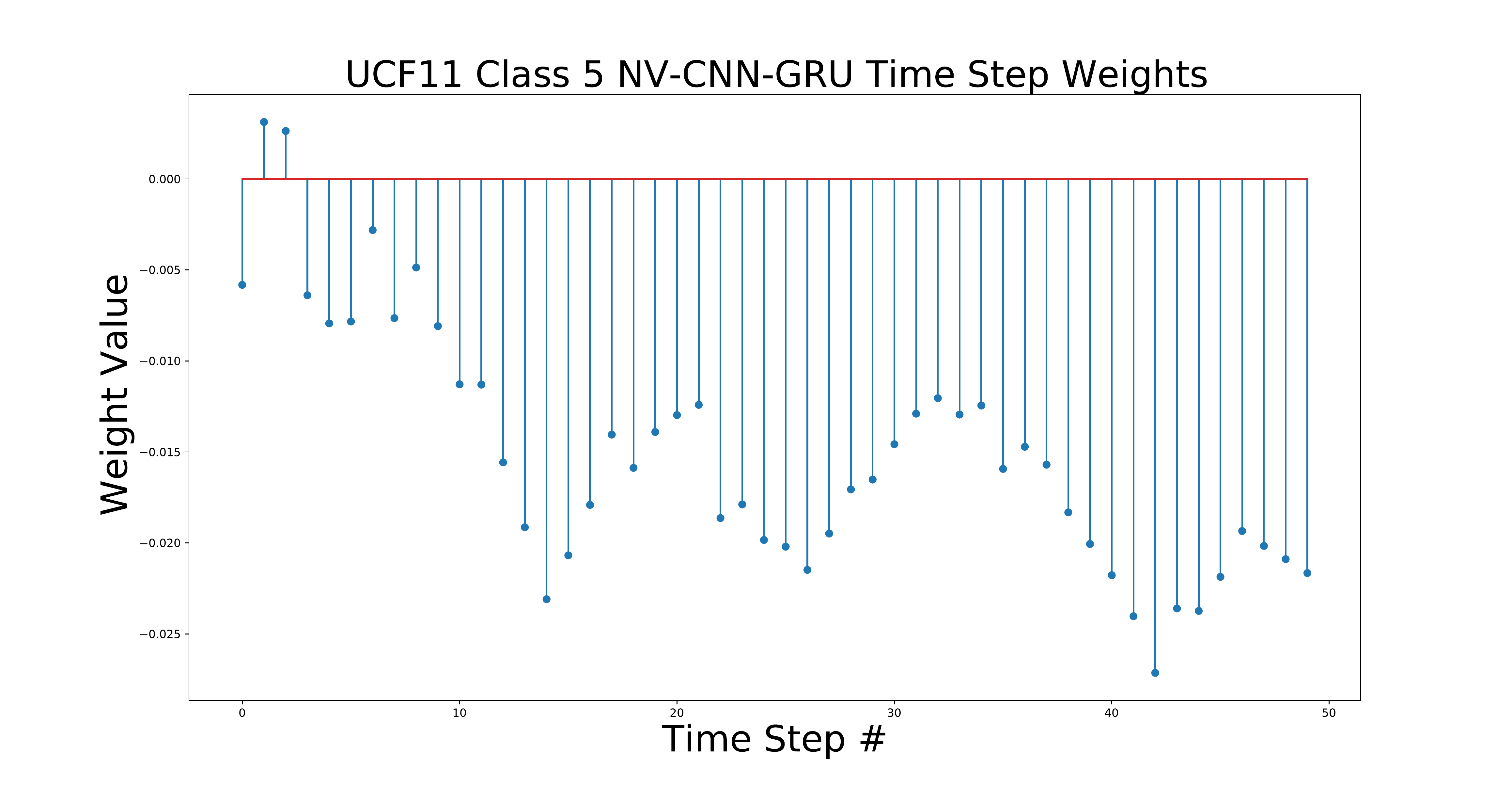}

\caption{Time step prioritization and filter unit prioritization (Class 5) for the NV-CNN-GRU model for UCF11. For this model, the positive mean weights are for the first layer of the CNN portion. Here the time step mean weights are negative.}
\label{fig:ucf11Class2}

\end{figure}

In Figure~\ref{fig:ucf11Class2}, we see that the mean positive weights are towards the CNN filter units as opposed to the GRU hidden states. This is interesting to notice that most of the hidden state time step mean weights are negative. This does make sense since the model is sequential and the NV-GRU is depending on the CNN's features. In addition, the dataset sampler is sampling 50 sequential frames and most of the videos contain more than 50 frames. Thus, the impression is that the hidden states may not be providing as much useful information as the CNN. From this work, \cite{li2018resound}, there is evidence that video action recognition datasets tend to have visual bias so it does make sense that with the NV-CNN-GRU model, it has more positive weights towards the spatial information. Given this information, there could be future work to mitigate the CNN/GRU prioritization while retaining the accuracy. Ideas from this work, \cite{nam2020learning}, could aid in mitigating the prioritization.

\begin{table}[h!]
    \centering
    \caption{Performance of the model with Time Analysis. From NV-GRU we inspect which time steps contribute  to each class by their weight value. In the Chinatown dataset, we remove the top weights for each class and calculate the overall class accuracy. Immediately, the test accuracy drops but then will come back up.}
    \begin{tabular}{|c|c|c|} \hline 
       Network  & \# of Time Steps & Test Accuracy  \\ \hline
       NV-GRU  & 0 & 96.4 \\ \hline
       NV-GRU  & 1 & 72.7 \\ \hline
       NV-GRU  & 5 & 43.87 \\ \hline
       NV-GRU  & 10 & 89.5 \\ \hline
    \end{tabular}
    
    \label{tab:testAccuracy}
\end{table}

\textbf{Counterfactuals}
There have been works \cite{chen2021human,poulos2021rnn,bica2020estimating,tonekaboni2019explaining} in using counterfactuals to understand how the RNN is performing. \cite{chen2021human,poulos2021rnn,bica2020estimating} use the counterfactuals for regression applications with RNNs. While \cite{tonekaboni2019explaining} uses the counterfactuals for classification applications with RNNs.

From the interpretability and explainability that NV-RNN provides, we can create counterfactuals called \textit{Time Analysis}. With each class having a unique set of weights per time step, we look at the mean hidden state weights per time step and set the top \textbf{K} time steps to zero. The idea is that the performance should drop if the information from the time-series data related to the time step is omitted. it will make it harder for the classifier to perform adequately. We set the top \textbf{K} time steps to zero to evaluate the degradation of the per-class accuracy.


For Time Analysis, we utilize the information we learned from Figure~\ref{fig:chinatownNV}, which shows that for each class there were about 5 top positive time steps. In Table~\ref{tab:testAccuracy}, this confirms that if we set those time steps to 0, then the class accuracy will drop. The interesting aspect is that after we remove more time steps, the class accuracy will increase but never get to the level of the original test accuracy. Note that this accuracy is for one of the NV-GRU models that did not achieve the best accuracy listed in Table~\ref{table:1}. The reason for this dip is that we are now eliminating the negative time steps so it will affect the classification decision. Additional experiments are in the Supplementary Material Section in \ref{counterfactualsSM} where there are individual class performance results.

\section{Conclusion}
 We propose a novel model, NV-RNN, as an alternative to traditional RNNs that has superior to on par performance to RNNs under multiple datasets. NV-RNN can provide interpretability and explain the prediction in a mathematical formulation. NV-RNN has the potential to be used in a wide array of different RNN applications. With its generic framework, it is used to show the connection between all of the hidden states and the classification. Thus, there are other scenarios within the scope of RNNs where it can provide additional understanding where typical RNNs are unable to provide. Overall, this type of interpretability and explainability is helpful to understand what is happening. In addition, with the concatenation of all the hidden states, it enables us to understand more and in most cases have higher performance. 
 

\begin{acks}

This work was supported by NSF grants CCF-1911094, IIS-1838177, and IIS-1730574; ONR grants N00014-18-12571, N00014-20-1-2534, and MURI N00014-20-1-2787; AFOSR grant FA9550-22-1-0060; and a Vannevar Bush Faculty Fellowship, ONR grant N00014-18-1-2047. We would like to thank Yehuda Dar, Hamid Javadi, Vishwanath Saragadam, and Fernando Gama for providing great comments and suggestions for this article.
\end{acks}


\bibliographystyle{ACM-Reference-Format}
\bibliography{sample-base}

\newpage
\appendix

\section{Supplementary Material}

\subsection{Architecture descriptions} \label{sm-arch}
In this section, we provide the recursive block computation of other recurrent architecture variants that has been used in this paper. To convert each model to it's NeuroView version, we concatenate all hidden states $\vh^{(t)}(\vx)$ and calculate the output using as described in Equation ~\ref{eq:VtQ}.

\subsubsection{Gated Recurrent Unit (GRU)}
\begin{align}
    \vr^{(t)}(\vx) &= \phi(\mW_{ir}\vx_t + \vb_{ir} + \mW_{hr}\vh^{(t-1)}(\vx) + \vb_{hr}) \\
    \vz^{(t)}(\vx) &= \phi(\mW_{iz}\vx_t + \vb_{iz} + \mW_{hz}\vh^{(t-1)}(\vx) + \vb_{hz}) \\
    \vn^{(t)}(\vx) &= \mathrm{tanh}( \mW_{in}\vx_t + \vb_{in} + \vr^{(t)}(\vx) \odot ( \mW_{hn}\vh^{(t-1)}(\vx) + \vb_{hn} ) ) \\
    \vh^{(t)}(\vx) &= (1 - \vz^{(t)}(\vx)) \odot \vn^{(t)}(\vx) + \vz^{(t)}(\vx) \odot \vz^{(t-1)}(\vx),
\end{align}
where $\phi(\alpha) = \frac{1}{1 + e^{( - \alpha)}}$ is the sigmoid function and $\odot$ is the Hadamard product.

\subsubsection{Long Short-Term Memory (LSTM) }
\begin{align}
    \vi^{(t)}(\vx) &= \phi(\mW_{ii}\vx_t + \vb_{ii} + \mW_{hi}\vh^{(t-1)}(\vx) + \vb_{hi}) \\
    \vf^{(t)}(\vx) &= \phi(\mW_{if}\vx_t + \vb_{if} + \mW_{hf}\vh^{(t-1)}(\vx) + \vb_{hf}) \\
    \vg^{(t)}(\vx) &= \mathrm{tanh}(\mW_{ig}\vx_t + \vb_{ig} + \mW_{hg}\vh^{(t-1)}(\vx) + \vb_{hg}) \\
    \vo^{(t)}(\vx) &= \phi(\mW_{io}\vx_t + \vb_{io} + \mW_{ho}\vh^{(t-1)}(\vx) + \vb_{ho})\\
    \vc^{(t)}(\vx) &= \vf^{(t)} \odot \vc^{(t-1)}(\vx) +  \vi^{(t)} \odot \vg^{(t)}(\vx) \\
    \vh^{(t)}(\vx) &= \vo^{(t)}(\vx) \odot \mathrm{tanh}(\vc^{(t)}(\vx)) 
\end{align}

\subsection{Hyperparameter Optimization}
\label{section:hyper}
From Table~\ref{table:1}, the results were taking the best model among different hidden states sizes which were 32, 64, and 128. The optimizer used is Adam. The learning rate is 0.001. The number of epochs is 1000.

For Large Movie Review dataset, we first preprocess the text by converting it to lowercase, removing all numbers, punctuation and special characters. Then for Word2Vec and Fasttext embedding method, we use dimensionality of 100 for word vectors. We count all words that appear at least once into our vocabulary, resulting in a vocabulary of size 122,762. Within the model training, the maximum distance between the current and predicted word within a sentence is 5. Both models are trained with the movie review dataset itself without any pre-training process. For GloVe embedding, we use a pre-trained model \textbf{GloVe.6B} whose embedding is trained on \textit{Wikipedia 2014} and \textit{Gigaword 5th Edition corpora} with 6 billion word tokens and 400,000 vocabulary size. We use the word embedding dimension of 100 so that it is consistent with other embedding methods. 

In training the models in Table~\ref{table:2}, the three different hidden state sizes used were 32, 64, and 128. The models were trained for 20 epochs. The optimizer used is Adam. The learning rate is 0.001. The batch size is 100.

In training the models in Table~\ref{table:3}, the hidden state size is set to 32. The number of epochs is 100. The learning rates for NV-CNN-RNN, NV-CNN-GRU, and NV-CNN-LSTM were set to 0.001 while the learning rates for CNN-RNN, CNN-GRU, and CNN-LSTM were set to 0.0001. The CNN architecture used was a 3 layer CNN with each CNN layer having a max-pooling layer placed after the CNN layer. The first CNN layer had an input of 3 channels and an output of 32 channels with a kernel size of 3 and padding of 1. The second CNN layer had an input of 32 channels and an output of 64 channels with a kernel size of 3 and a padding of 1. The third CNN layer had an input of 64 channels and an output of 64 channels with a kernel size of 3 and padding of 1. The activation function used is a ReLU. Every max pooling parameter would pool by a factor of 2. The input size for all the RNNs, GRUs, and LSTMs would be 28*28*64.

\subsection{Ablation Studies}
\label{section:ablation}
We conduct ablation studies for different hidden state dimension of 32, 64, 128 for all the models mentioned in Table \ref{table:1}. The results can be found in Table \ref{tab:abla-nv}, Table \ref{tab:abla-base} and Table \ref{tab:abla-avg}.

\begin{table}[h]
\centering
\caption{An ablation study of NV-RNN, NV-GRU, and NV-LSTM models with hidden state sizes of 32, 64, 128.}
{
\begin{adjustbox}{width=\linewidth,center}
\begin{tabular}{|l|ccc|ccc|ccc|}
\toprule
\textbf{Data set}        & \textbf{NV-GRU32} & \textbf{NV-GRU64} &\textbf{NV-GRU128} & \textbf{NV-RNN32} & \textbf{NV-RNN64} & \textbf{NV-RNN128} & \textbf{NV-LSTM32} & \textbf{NV-LSTM64} & \textbf{NV-LSTM128} \\ \hline 
Adiac  & \textbf{68.28\%} & 64.45\% & 68.03\% & 68.79\% & 64.96\% & \textbf{69.56\%} & 68.03\% & 71.35\% & \textbf{74.68\%} \\
BME & \textbf{98.66\%} & \textbf{98.66\%} & \textbf{98.66\%} & 98.66\% & \textbf{99.33\%} & 98.66\% & \textbf{98.66\%} & 98\%  & \textbf{98.66\%} \\
CBF & 95.66\% & 98\% & \textbf{98.44\%} & \textbf{97.77\%} & 96\% & 97.44\% & 98.11\% & 98.22\% & \textbf{98.55\%} \\
Chinatown & 96.5\% & \textbf{97.08\%} & \textbf{97.08\%} & \textbf{97.95\%} & 97.08\% & \textbf{97.95\%} & \textbf{98.54\%} & \textbf{98.54\%} & 98.25\% \\
Chlorine Concentration & 74.58\% & 75.44\% & \textbf{78.15\%} & 80.33\% & 82.21\% & \textbf{83.95\%} & 68.82\% & 69.01\% & \textbf{72.39}\% \\
Fungi & \textbf{98.92\%} & 97.84\% & 96.23\% & \textbf{96.77\%} & 96.23\% & 94.08\% & 98.92\% & \textbf{99.46\%} & 98.38\% \\
Ham & \textbf{80.95\%} & 78.09\% & 79.04\% & \textbf{78.09\%} & 75.23\% & \textbf{78.09\%} & 77.14\% & 77.14\% & \textbf{78.09\%} \\
Haptics & \textbf{46.1\%} & 45.77\% & 45.45\% & \textbf{47.07\%} & 45.12\% & 46.42\% & 45.45\% & \textbf{45.77\%} & 44.8\% \\
Herring & 71.87\% & \textbf{73.43\%} & 68.75\% & \textbf{68.75\%} & \textbf{68.75\%} & \textbf{68.75\%} & \textbf{68.75\%} & \textbf{68.75\%} & 67.18\% \\
InsectRT & \textbf{100\%} & \textbf{100\%} & \textbf{100\%} & \textbf{100\%} & \textbf{100\%} & \textbf{100\%} & \textbf{100\%} & \textbf{100\%} & \textbf{100\%} \\
InsectST & \textbf{100\%} & \textbf{100\%} & \textbf{100\%} & \textbf{100\%} & \textbf{100\%} & \textbf{100\%} & \textbf{100\%} & \textbf{100\%} & \textbf{100\%} \\
InsectWingbeat & \textbf{64.54\%} & 64.39\% & 64.04\% & 63.98\% & 64.19\% & \textbf{64.29\%} & 63.58\% & \textbf{63.88\%} & 63.73\% \\
Meat & 91.66\% & \textbf{96.66\%} & 95\% & 95\% & 93.33\% & \textbf{96.66\%} & \textbf{96.66\%} & 93.33\% & 93.33\% \\
OliveOil & \textbf{93.33\%} & \textbf{93.33\%} & 90\% & 90\% & \textbf{93.33\%} & \textbf{93.33\%} & 90\% & \textbf{93.33\%} & \textbf{93.33\%} \\
Plane & 98.09\% & 98.09\% & \textbf{100\%} & \textbf{99.04\%} & 98.09\% & 97.14\% & 97.14\% & \textbf{99.04\%} & \textbf{99.04\%} \\
Rock & 68\% & 70\% & \textbf{76\%} & 76\% & \textbf{80\%} & 72\% & \textbf{82\%} & 74\% & 80\% \\
SmoothSubspace & \textbf{96\%} & 94.66\% & 95.33\% & 90.66\% & \textbf{91.33\%} & 90.66\% & \textbf{94\%} & 90\% & 90\% \\
Synthetic Control & \textbf{99.33\%} & 98.66\% & 98.66\% & 99.33\% & 98.33\% & \textbf{99.66\%} & \textbf{99.33\%} & 98.33\% & 99\% \\
UMD & \textbf{100\%} & \textbf{100\%} & \textbf{100\%} & \textbf{100\%} & \textbf{100\%} & \textbf{100\%} & \textbf{100\%} & \textbf{100\%} & \textbf{100\%} \\
Wine & \textbf{100\%} & 98.14\% & \textbf{100\%} & \textbf{100\%} & \textbf{100\%} & \textbf{100\%} & \textbf{100\%} & \textbf{100\%} & 96.29\% \\

\bottomrule
\end{tabular}
\label{tab:abla-nv}
\end{adjustbox}
}
\label{tab:real-data-svm}
\end{table}

\begin{table}[h]
\centering
\caption{An ablation study of RNN, GRU, and LSTM models with hidden state sizes of 32, 64, 128.}
{
\begin{adjustbox}{width=\linewidth,center}
\begin{tabular}{|l|ccc|ccc|ccc|}
\toprule
\textbf{Data set}        & \textbf{GRU32} & \textbf{GRU64} &\textbf{GRU128} & \textbf{RNN32} & \textbf{RNN64} & \textbf{RNN128} & \textbf{LSTM32} & \textbf{LSTM64} & \textbf{LSTM128} \\ \hline 
Adiac  & 30.69\% & 36.82\% & \textbf{37.08\%} & 31.2\% & \textbf{35.8\%} & 33.75\% & 39.13\% & \textbf{49.61\%} & 48.33\% \\
BME & 93.33\% & \textbf{94.66\%} & 92.66\% & 77.33\% & \textbf{88.66\%} & 74\% & \textbf{80\%} & 76.66\%  & 78\% \\
CBF & 76.44\% & 81.66\% & \textbf{94.55\%} & \textbf{60.66\%} & 56.66\% & 57.22\% & \textbf{90\%} & 87.11\% & 84.55\% \\
Chinatown & 96.79\% & \textbf{97.37\%} & \textbf{97.37\%} & \textbf{74.34\%} & 72.59\% & 72.01\% & \textbf{97.66\%} & \textbf{97.66\%} & \textbf{97.66\%} \\
Chlorine Concentration & 58.07\% & \textbf{60.1\%} & 59.74\% & 56.48\% & 57.16\% & \textbf{58.17\%} & 56.71\% & 57.31\% & \textbf{57.73\%} \\
Fungi & 43.54\% & 50.53\% & \textbf{58.6\%} & \textbf{49.46\%} & 47.31\% & 46.23\% & 45.16\% & 67.2\% & \textbf{68.81\%} \\
Ham & \textbf{68.57\%} & 67.61\% & \textbf{68.57\%} & 68.57\% & \textbf{69.52\%} & \textbf{69.52\%} & \textbf{69.52\%} & \textbf{69.52\%} & 67.61\% \\
Haptics & 41.23\% & \textbf{41.55\%} & 40.25\% & 37.33\% & 40.58\% & \textbf{42.2\%} & 38.31\% & 41.23\% & \textbf{41.88\%} \\
Herring & 65.62\% & 65.62\% & \textbf{67.18\%} & 65.62\% & \textbf{67.18\%} & \textbf{67.18\%} & \textbf{68.75\%} & 65.62\% & 67.18\% \\
InsectRT & \textbf{100\%} & \textbf{100\%} & \textbf{100\%} & \textbf{100\%} & \textbf{100\%} & \textbf{100\%} & \textbf{100\%} & \textbf{100\%} & \textbf{100\%} \\
InsectST & \textbf{100\%} & \textbf{100\%} & \textbf{100\%} & \textbf{100\%} & \textbf{100\%} & \textbf{100\%} & \textbf{100\%} & \textbf{100\%} & \textbf{100\%} \\
InsectWingbeat & 44.34\% & 48.98\% & \textbf{49.34\%} & 26.61\% & \textbf{28.93}\% & 27.87\% & 38.68\% & \textbf{43.58\%} & 28.73\% \\
Meat & 45\% & 45\% & \textbf{50\%} & \textbf{48.33\%} & \textbf{48.33\%} & 45\% & \textbf{50\%} & 46.66\% & 41.66\% \\
OliveOil & \textbf{50\%} & 40\% & 40\% & \textbf{46.66\%} & 40\% & 40\% & \textbf{40\%} & \textbf{40\%}& \textbf{40\%} \\
Plane & 68.57\% & 67.61\% & \textbf{79.04\%} & 59.04\% & 62.85\% & \textbf{89.52\%} & 87.61\% & 91.42\% & \textbf{95.23\%} \\
Rock & 68\% & \textbf{74\%} & 64\% & \textbf{64\%} & 62\% & 60\% & 62\% & \textbf{68\%} & 66\% \\
SmoothSubspace & \textbf{89.33\%} & \textbf{89.33\%} & \textbf{89.33\%} & \textbf{91.33\%} & 90.66\% & 90.66\% & 88.66\% & 90\% & \textbf{90.66\%} \\
Synthetic Control & 98\% & \textbf{98.66\%} & 97.33\% & 79\% & 98\% & \textbf{99.66\%} & 96.33\% & \textbf{98.33\%} & \textbf{98.33\%} \\
UMD & 66.66\% & 88.88\% & \textbf{99.3\%} &\textbf{74.3\%} & 65.97\% & 66.66\% & 65.27\% & \textbf{86.8\%} & 67.36\% \\
Wine & \textbf{59.25\%} & 55.55\% & 55.55\% & 57.4\% & \textbf{59.25\%} & 50\% & \textbf{62.96\%} & 53.7\% & 59.25\% \\

\bottomrule
\end{tabular}
\label{tab:abla-base}
\end{adjustbox}
}
\label{tab:real-data-svm}
\end{table}

\begin{table}[h]
\centering
\caption{An ablation study of RNN-AVG, GRU-AVG, and LSTM-AVG models with hidden state sizes of 32, 64, 128.}
{
\begin{adjustbox}{width=\linewidth,center}
\begin{tabular}{|l|ccc|ccc|ccc|}
\toprule
\textbf{Data set}        & \textbf{GRU-AVG32} & \textbf{GRU-AVG64} &\textbf{GRU-AVG128} & \textbf{RNN-AVG32} & \textbf{RNN-AVG64} & \textbf{RNN-AVG128} & \textbf{LSTM-AVG32} & \textbf{LSTM-AVG64} & \textbf{LSTM-AVG128} \\ \hline 
Adiac  & 15.85\% & 19.18\% & \textbf{31.45\%} & \textbf{10.23\%} & 9.46\% & 7.92\% & 7.67\% & \textbf{16.87\%} & 9.2\% \\
BME & 84\% & 84\% & \textbf{84.66\%} & \textbf{84\%} & 72\% & \textbf{81.33\%} & 64\% & \textbf{84.66\%} & 84\% \\
CBF & 98.33\% & \textbf{98.66\%} & 97.66\% & \textbf{97.33\%} & 96.22\% & 96.66\% & 98\% & \textbf{99.77\%} & 97.44\% \\
Chinatown & \textbf{98.25\%} & 97.66\% & \textbf{98.25\%} & 98.54\% & \textbf{98.83\%} & \textbf{98.83\%} & \textbf{98.54\%} & \textbf{98.54\%} & \textbf{98.54\%} \\
Chlorine Concentration & 56.38\% & 56.87\% & \textbf{57.05\%} & 55.39\% & \textbf{55.44\%} & 55.07\% & 55.65\% & \textbf{55.88\%} & 55.2\% \\
Fungi & 39.78\% & 48.92\% & \textbf{58.6\%} &  40.86\% & 46.23\% & \textbf{60.21\%} & 58.6\% & 58.06\% & \textbf{75.26\%} \\
Ham & 75.23\% & \textbf{81.9\%} & 77.14\% & \textbf{74.28\%} & 73.33\% & 71.42\% & \textbf{80.95\%} & 75.23\% & 77.14\% \\
Haptics & 38.63\% & 42.53\% & \textbf{44.48\%} & 32.79\% & 33.11\% & \textbf{33.76\%} & 34.74\% & \textbf{41.88}\% & 39.61\% \\
Herring & 62.5\% & \textbf{65.62\%} & \textbf{65.62\%} & \textbf{67.18\%} & 64.06\% & 59.37\% & 60.93\% & 64.06\% & \textbf{68.75\%} \\
InsectRT & \textbf{100\%} & \textbf{100\%} & \textbf{100\%} & \textbf{100\%} & \textbf{100\%} & \textbf{100\%} & \textbf{100\%} & \textbf{100\%} & \textbf{100\%} \\
InsectST & \textbf{100\%} & \textbf{100\%} & \textbf{100\%} & \textbf{100\%} & \textbf{100\%} & \textbf{100\%} & \textbf{100\%} & \textbf{100\%} & \textbf{100\%} \\
InsectWingbeat & 26.26\% & \textbf{46.41\%} & 44.69\% & \textbf{28.48}\% & 23.03\% & 26.16\% & 32.47\% & 32.72\% & \textbf{39.94\%} \\
Meat & 66.66\% & 68.33\% & \textbf{86.66\%} & \textbf{66.66\%} & \textbf{66.66\%} & 65\% & \textbf{81.66\%} & 66.66\% & 65\% \\
OliveOil & 40\% & 40\% & \textbf{80\%} & \textbf{40\%} & \textbf{40\%} & \textbf{40\%} & \textbf{40\%} & \textbf{40\%} & \textbf{40\%} \\
Plane & 65.71\% & 93.33\% & \textbf{98.09\%} & \textbf{65.71\%} & 59.04\% & 62.85\% & 69.52\% & \textbf{70.47\%} & 68.57\% \\
Rock & 52\% & 54\% & \textbf{62\%} & 52\% & \textbf{56\%} & 50\% & 50\% & 56\% & \textbf{60\%} \\
SmoothSubspace & \textbf{91.33\%} & 88\% & 88\% & 89.33\% & 90\% & \textbf{90.66\%} & \textbf{86.66\%} & 84\% & 86\% \\
Synthetic Control & 94.66\% & 92.33\% & \textbf{95.66\%} & 88\% & 85.33\% & \textbf{94.33\%} & 96\% & 95.66\% & \textbf{97.33\%} \\
UMD & 66.66\% & 84.72\% & \textbf{92.36\%} & 66.66\% & \textbf{75\%} & 72.91\% & 70.83\% & \textbf{72.22\%} & \textbf{72.22\%} \\
Wine & \textbf{79.62\%} & 72.22\% & 66.66\% & \textbf{75.92\%} & 61.11\% & 66.66\% & \textbf{74.07\%} & 72.22\% & 66.66\% \\

\bottomrule
\end{tabular}
\label{tab:abla-avg}
\end{adjustbox}
}
\label{tab:real-data-svm}
\end{table}

\subsection{Additional Counterfactuals}
\label{counterfactualsSM}

Table~\ref{tab:TimeAttack} shows the results of using Time Analysis on the test dataset. By focusing on one class at a time, we can quantitatively assess how perturbing the time steps within the data can affect the performance. With one time step modified, the accuracy dropped by one percent. When we started to set more of the input data at certain time steps to zero, the accuracy would continue to drop.

One interesting observation is to see which time steps have a negative mean value and to set the most negative weights to zero. Table~\ref{tab:TimeAttackNeg} displays the test accuracy for the negative Time Analysis. The test accuracy will remain the same, but when ten input data time steps are set to zero the test accuracy of that class has increased. Hence the interpretability of this model can provide these insights and practitioners can develop test cases for their data to understand what happens if you perturb the data to certain degrees. 

\begin{table}[h!]
    \centering
    \caption{Performance of the model with Time Analysis with the Insect Wingbeat dataset. From NV-LSTM we inspect which time steps correspond to the most significant positive weights. From there we set the input data at those time steps to zero. As we set more of the time indices to zero, the test performance will start to decrease more and more. }
    \begin{tabular}{|c|c|c|c|} \hline 
       Network & Class  & \# of Time Steps & Per Class Accuracy  \\ \hline
       NV-LSTM & 0th  & 0 & 81.1 \\ \hline
       NV-LSTM & 0th & 1 & 80.0 \\ \hline
       NV-LSTM & 0th & 5 & 77.2 \\ \hline
       NV-LSTM & 0th & 10 & 70.0 \\ \hline
       NV-LSTM & 0th & 15 & 70.0 \\ \hline
       NV-LSTM & 0th & 20 & 68.3 \\ \hline
       NV-LSTM & 0th & 25 & 65.5 \\ \hline
       NV-LSTM & 0th & 30 & 61.1 \\ \hline 
    \end{tabular}
    
    \label{tab:TimeAttack}
\end{table}

\begin{table}[h!]
    \centering
    \caption{Performance of the model with Time Analysis with the Insect Wingbeat dataset. From NV-LSTM we inspect which time steps correspond to negative weights we can set the input data to those indices to zero. When we set at least 10 of them, we see that the class's test accuracy increases. From the previous table, setting the time indices from the positive weights would decrease the performance, but here we see the opposite effect.}
    \begin{tabular}{|c|c|c|c|} \hline 
       Network & Class  & \# of Time Steps & Per Class Accuracy  \\ \hline
       NV-LSTM & 0th  & 0 & 81.1 \\ \hline
       NV-LSTM & 0th & 1 & 81.1 \\ \hline
       NV-LSTM & 0th & 5 & 81.1 \\ \hline
       NV-LSTM & 0th & 10 & 82.2 \\ \hline
    \end{tabular}
    
    \label{tab:TimeAttackNeg}
\end{table}

\end{document}